\documentclass{article} 
\usepackage{iclr2027_conference,times}

\usepackage{hyperref}
\usepackage{url}
\usepackage{graphicx}
\usepackage{booktabs}
\usepackage{amsmath}
\usepackage{amssymb}
\usepackage{multirow}
\usepackage{algorithm}
\usepackage{algpseudocode}
\usepackage{enumitem}
\usepackage{subcaption}
\usepackage{caption}
\usepackage{tikz}          
\usepackage[T1]{fontenc}
\usepackage{xcolor}
\usepackage{upquote}
\usepackage{tcolorbox}
\usepackage{tabularx}
\tcbuselibrary{skins,breakable,listings}
\usepackage[table]{xcolor}
\usepackage{titletoc}

\definecolor{promptframe}{RGB}{55,45,72}
\definecolor{promptback}{RGB}{246,246,246}

\definecolor{promptback}{HTML}{FAFBFC}
\definecolor{promptframe}{HTML}{B8C4D2}
\definecolor{prompttitleback}{HTML}{EDF2F7}
\definecolor{prompttitle}{HTML}{294866}

\newtcblisting{promptbox}[1]{
    enhanced,
    breakable,
    listing only,
    listing engine=listings,
    width=\linewidth,
    colback=promptback,
    colframe=promptframe,
    colbacktitle=prompttitleback,
    coltitle=prompttitle,
    fonttitle=\normalfont\footnotesize\bfseries,
    title={#1},
    boxrule=0.4pt,
    titlerule=0pt,
    arc=1mm,
    outer arc=1mm,
    boxsep=0pt,
    left=3mm,
    right=3mm,
    top=2mm,
    bottom=2mm,
    toptitle=1.5mm,
    bottomtitle=1.5mm,
    before skip=8pt,
    after skip=8pt,
    listing options={
        language={},
        basicstyle=\ttfamily\fontsize{7.8}{9.2}\selectfont,
        columns=fullflexible,
        keepspaces=true,
        breaklines=true,
        breakatwhitespace=true,
        breakindent=1em,
        showstringspaces=false,
        showspaces=false,
        showtabs=false,
        upquote=true,
        tabsize=2,
        numbers=none,
        mathescape=false,
        texcl=false,
        aboveskip=0pt,
        belowskip=0pt,
        literate={—}{{\textemdash}}1
    }
}

\iclrfinalcopy

\title{Learning What to Remember: Long-horizon Counterfactual Memory Optimization}

\author{
Jiaming Tang$^{1}$, Mingyan Liu$^{1}$, Armin Sarabi$^{1}$ \\
\\
$^{1}$Department of Electrical Engineering and Computer Science, University of Michigan
}

\newcommand{\appref}[1]{\hyperref[#1]{Appendix~\ref*{#1}}}

\begin{document}

\maketitle

\fancyhead{}
\renewcommand{\headrulewidth}{0pt}
\pagestyle{fancy}

\begin{abstract}
Persistent textual memory allows language models to carry information across long interactions, but learning what to remember is fundamentally a credit-assignment problem. A memory rewrite may only become useful many steps later, while much of the observed utility may be inherited from information already stored before the rewrite. We introduce Memory Gain Policy Optimization (MGPO), which isolates the incremental value of each memory rewrite by crediting it for its marginal contribution to current and future downstream utility. This turns delayed memory utility into a direct learning signal for optimizing what information should persist. We study MGPO on document-level information extraction, where structured supervision makes the effects of individual memory updates directly measurable. MGPO improves extraction while reducing average memory length by nearly 80\% relative to the initial memory policy before optimization. The learned memory policy also supports reuse and transfer across domains, downstream models without further training. These results show that effective memory learning depends not only on preserving useful information, but on identifying which memory updates create lasting incremental value.
\end{abstract}

\section{Introduction}

Persistent memory allows language models to maintain information across interactions, but limited capacity requires them to selectively retain, revise, or discard stored content \citep{toshniwal2020learning, wang2023augmenting, yu2026memagent, zhou2026mem1}. Learning such update decisions is challenging because information that appears irrelevant at the time of writing may become useful only much later \citep{arjona2019rudder, han2019episodic, hung2019optimizing}. Moreover, even when a memory proves useful for a downstream task, its utility may be largely attributable to information inherited from the preceding memory state rather than to the most recent revision. As a result, the utility of the resulting memory does not by itself reveal how much the latest update contributed \citep{setlur2025rewarding}. Learning what to remember thus means learning not only whether a memory update is useful now, but also its incremental value for the future beyond what the memory already knows.

Existing memory systems support selective retention through retrieval \citep{guo2025lightrag}, explicit memory management \citep{chhikara2025mem0, xu2026mem, yu2026agentic}, and learned policies that construct and revise persistent memory \citep{yu2026memagent, yan2026himpo, zhou2026mem1}. Yet these approaches largely leave a central learning problem unresolved. It remains unclear whether a memory update actually creates downstream value. To make the effect of memory updates measurable, we optimize a persistent memory writer against a fixed downstream reader, following prior work on upstream adaptation against frozen downstream models \citep{xu2024recomp, yang2023prca, yoon2024compact}. This leads to a more fundamental credit-assignment problem: when an updated memory improves downstream utility, how much of that utility is created by the latest rewrite, rather than inherited from the memory that already existed?

To answer this question, consider a reader working through a long document with a small notebook. A natural way to assess a notebook revision is to retain both versions and compare how well the same reader performs on the same downstream tasks using each. Information already present contributes to both evaluations, so their difference isolates the change in utility attributable to the revision, following the broader principle \citep{tumer2002learning, harutyunyan2019hindsight, foerster2018counterfactual} of counterfactual credit assignment. Persistent memory also has value through future reuse. A recorded detail may offer little immediate benefit yet help interpret subsequent evidence after the original passage has left the reader’s context. This motivates long-horizon counterfactual credit assignment, where the pre- and post-rewrite memories are evaluated with the same reader and downstream target, and the paired comparison is propagated across subsequent targets that reuse the updated memory.

We propose the Memory Gain Policy Optimization (MGPO), which trains the memory writer using counterfactual utility differences measured by the frozen reader across current and future downstream targets. MGPO uses these differences to credit each rewrite for its current and future utility, with policy-gradient returns accounting for subsequent rewrites along the memory trajectory. We show that the resulting return decomposes into the cumulative downstream utility minus the inherited utility of the pre-rewrite memory over the same remaining horizon. Because this inherited component is independent of the current rewrite, MGPO preserves the expected policy gradient of the underlying factual objective while changing how credit is assigned to individual memory rewrites.

We evaluate MGPO on document-level information extraction (IE), where predictions often depend on evidence distributed across distant passages \citep{yao2019docred, tan2022revisiting, jain2020scirex}, providing a natural setting for studying the value of memory through later reuse. Our experiments demonstrate that MGPO learns compact, effective memories that support the integration of evidence across distant passages and improve downstream extraction performance. The learned writer also transfers across unseen readers, domains, and downstream tasks, showing that the benefits of memory construction extend beyond the reader and task used during training.

The main contributions of this paper are as follows:
\begin{itemize}[leftmargin=*, noitemsep]
    \item We identify inherited utility as a central challenge in persistent memory learning. The utility of a memory state may largely reflect previously stored information, obscuring the contribution of individual memory rewrites.
    \item We introduce Memory Gain Policy Optimization (MGPO), which assigns each rewrite long-horizon counterfactual credit. MGPO preserves the original downstream objective while providing a more informative learning signal for memory updates.
    \item We show that MGPO learns compact and selective memories, improves document-level information extraction, and generalizes beyond its training setting through reuse across readers and transfer across downstream domains.
\end{itemize}

\section{Methodology}
\label{sec:method}

\begin{figure} [t]
    \centering
    \includegraphics[width=0.9\linewidth]{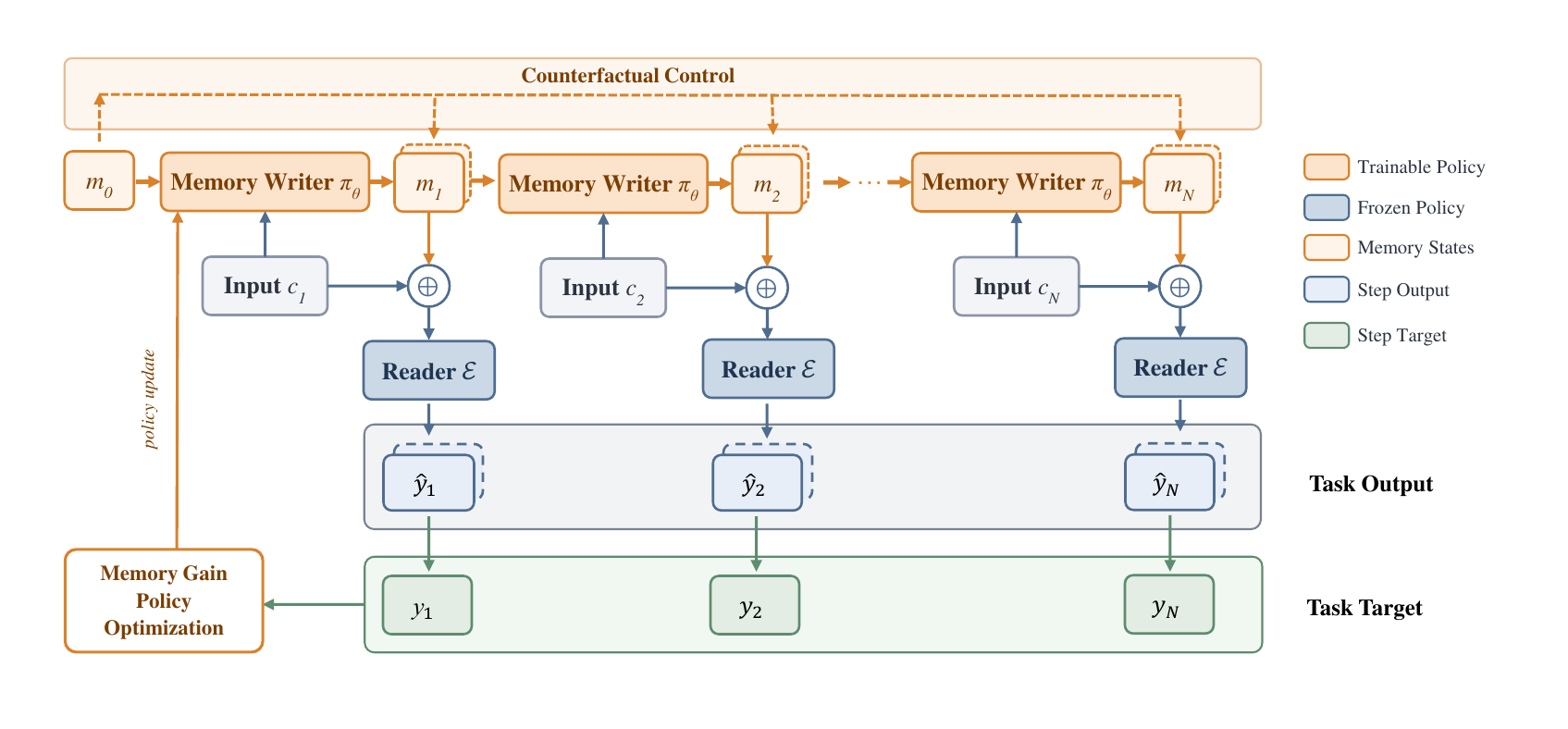}
    \caption{Overview of our methodology. A trainable writer sequentially updates a persistent memory state, while a frozen reader evaluates its downstream utility. A counterfactual comparison evaluates the pre- and post-rewrite memories on the same current and future targets. Summing these utility differences yields the \emph{Memory Gain}, which serves as the learning signal for optimizing the writer.}
    \label{fig:workflow}
\end{figure}

Persistent-memory learning faces a fundamental attribution ambiguity: state utility conflates what is inherited from prior memory with what the current rewrite contributes. MGPO isolates the latter by counterfactually comparing adjacent memory states over their downstream reuse horizon. The resulting credit is exactly potential-shaped, preserving the factual policy objective while reallocating learning signal to the rewrites that generate downstream utility.

\subsection{Decoupled persistent memory writing}
\label{sec:arch}

As illustrated in \autoref{fig:workflow}, we consider a streaming process where inputs $c_1,\ldots,c_N$ arrive sequentially, with only a bounded persistent memory carried forward. At step $t$, a learnable memory writer $\pi_\theta$ processes the previous memory and new input to generate an updated memory state:
\begin{equation}
m_t \sim \pi_\theta(\cdot \mid m_{t-1}, c_t),
\qquad
m_0 = \varnothing.
\qquad
|m_t| \le B.
\end{equation}
The resulting memory text replaces the previous memory under a fixed budget $B$, so optimizing the writer entails learning what information should be retained, discarded, or reorganized as the state evolves. Crucially, we decouple this writing process from answer generation: the writer $\pi_\theta$ remains the sole trainable component, while a fixed reader $\mathcal{E}$ processes the memory state $m_t$ together with downstream input $c_j$ to yield an answer:
\begin{equation} 
\hat{y}_{t,j}(\xi)=\mathcal{E}(m_t,c_j;\xi). 
\end{equation}
Here, $\xi$ denotes reader randomness. We then define the utility of memory state $m_t$ on target $j$ as:
\begin{equation}
F_{t,j} = \mathbb{E}_{\xi}\left[\mathcal{U}\left(\hat y_{t,j}(\xi), y_j\right)\right],
\end{equation}
where $y_j$ denotes the target associated with $c_j$ and $\mathcal{U}$ is a task-specific utility function. The expectation is taken over the fixed sampling distribution of the downstream reader. In practice, we estimate each $F_{t,j}$ with one Monte Carlo sample from the same fixed reader decoding distribution. An empirical analysis of sampling noise are reported in \appref{app:reader_noise}. This separation makes memory independently measurable. Because the reader, target, and utility function are fixed, differences in $F_{t,j}$ isolate the effect of the supplied memory state, enabling counterfactual evaluation of individual rewrites. 

\subsection{From memory-state utility to rewrite credit}
\label{sec:credit_assign}

The quantity $F_{t,j}$ evaluates a memory state, but policy optimization requires credit for the action that produced that state. These are not equivalent: a high $F_{t,j}$ may largely reflect information that was already present in $m_{t-1}$ rather than from rewrite $t$ itself. We refer to this pre-existing contribution as \emph{inherited utility}. To attribute utility to the rewrite itself, we evaluate the pre- and post-rewrite memories on the same downstream target:
\begin{equation}
\Delta_{t,j}
=
F_{t,j}-F_{t-1,j}.
\end{equation}
Because all downstream factors are held fixed, $\Delta_{t,j}$ measures the change in utility associated with the rewrite $t$ on target $j$.

\begin{figure} [t]
    \centering
    \includegraphics[width=0.8\linewidth]{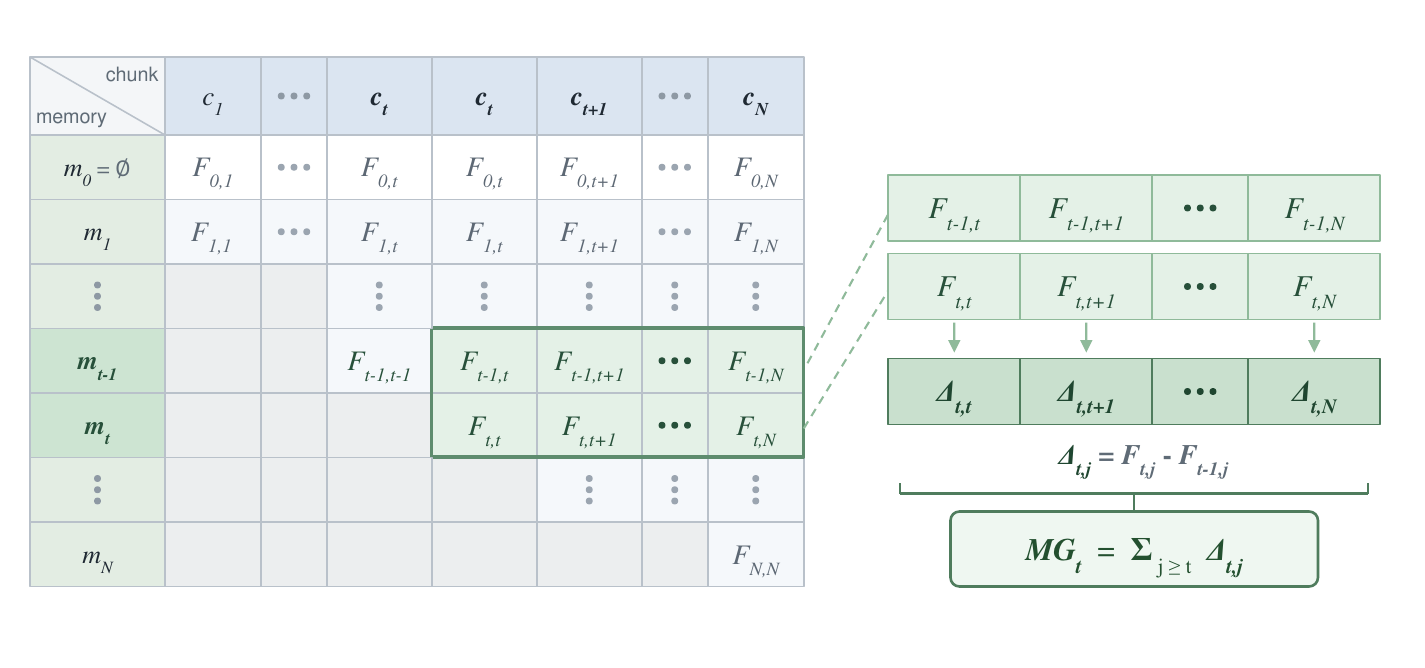}
    \caption{Counterfactual evaluation matrix. $F_{t,j}$ measures the utility of memory state $m_t$ on downstream target $j$, with the diagonal corresponding to the streaming trajectory during evaluation. The difference $F_{t,j}-F_{t-1,j}$ attributes the marginal utility on target $j$ to rewrite $t$, and $MG_t$ aggregates this contribution over all current and future targets ($j\geq t$).}
    \label{fig:cf_matrix}
\end{figure}

A rewrite can matter beyond the prediction immediately after it. Information introduced at step $t$ may become useful only when combined with evidence encountered several steps later. We therefore evaluate the same rewrite over its entire remaining reuse horizon and define its \emph{Memory Gain} as:
\begin{equation}
MG_t
=
\sum_{j=t}^{N}\Delta_{t,j}
=
\sum_{j=t}^{N}
\left(
F_{t,j}-F_{t-1,j}
\right).
\end{equation}
The quantities $F_{t,j}$ form the upper-triangular counterfactual matrix in \autoref{fig:cf_matrix}. The diagonal $F_{j,j}$ corresponds to the factual streaming trajectory, whereas off-diagonal entries evaluate how an earlier memory state would support later targets. These counterfactual evaluations are needed only during training, while inference follows the diagonal factual trajectory. \appref{app:complexity} compares the training cost with group-based RL methods.

$MG_t$ aggregates the marginal effect of the transition from $m_{t-1}$ to $m_t$ over all current and future targets. Since $m_t$ also influences subsequent rewrites, we define the return for rewrite \(t\) as $G_t = \sum_{k=t}^{N}MG_k$, thereby including their future rewards. Thus, \(MG_t\) aggregates over downstream targets for a single rewrite, whereas \(G_t\) aggregates over the current and subsequent rewrites.

\subsection{Memory Gain as potential-shaped credit}

Memory Gain attributes downstream utility to individual memory rewrites. We now show that this rewrite-level credit remains aligned with the original objective along the factual streaming trajectory. 

Along the factual trajectory, the memory produced at step $t$ is evaluated on its corresponding target, giving the per-step reward $r_t^{\mathrm{fact}} = F_{t,t},$ and the remaining factual return $G^{\text{fact}}_t=\sum^N_{k=t}r_k^{\text{fact}}$. Unlike \(r_t^{\mathrm{fact}}\), which evaluates the utility of the resulting memory state, \(MG_t\) measures the marginal contribution of rewrite \(t\) across its current and future reuse. \(G_t\) additionally accounts for rewards assigned to subsequent rewrites. To relate this return to the factual objective, we define the inherited utility already supported by the pre-rewrite memory over the remaining horizon as:
\begin{equation}
\Phi_t
=
\sum_{j=t}^{N} F_{t-1,j},
\qquad
\Phi_{N+1}=0.
\end{equation}
By the definition of Memory Gain, we have:
\begin{equation}
MG_t
=
r_t^{\mathrm{fact}}
+
\Phi_{t+1}
-
\Phi_t.
\end{equation}
Hence, Memory Gain is a potential-shaped form of the factual streaming reward. Subtracting $\Phi_t$ removes inherited utility from preceding memory, while adding $\Phi_{t+1}$ captures the updated memory's future utility, assigning credit according to each rewrite's incremental downstream contribution.

Summing over the remaining trajectory telescopes the potential terms:
\begin{equation}
G_t
=
G_t^{\mathrm{fact}}
-
\Phi_t.
\end{equation}
We next show that this preserves the expected policy gradient of the underlying objective while changing its finite-sample gradient estimator. Let
\begin{equation}
    \psi_t = \nabla_\theta \log \pi_\theta \!\left( m_t \mid m_{t-1}, c_t \right).
\end{equation}
Then the Memory Gain gradient estimator at step \(t\) can be written as
\begin{equation}
    \hat g_t =  \psi_t G_t =  \hat g_t^{\mathrm{fact}} - \psi_t \Phi_t.
\end{equation}
Since \(\Phi_t\) is determined entirely by the pre-rewrite memory \(m_{t-1}\) and the remaining targets, it is determined before \(m_t\) is sampled and is therefore independent of the current action conditional on the pre-rewrite history. Therefore, $\mathbb{E}[\psi_t\Phi_t]=0$, and consequently:
\begin{equation}
\nabla_\theta J_{\mathrm{MG}}
=
\nabla_\theta J_{\mathrm{fact}}.
\end{equation}
MGPO therefore provides a different finite-sample estimator of the same factual policy gradient; empirically, we show in \autoref{sec:long_horizon_credit} that this estimator outperforms the estimator obtained by directly using the factual return. The counterfactual term \(\Phi_t\) acts as an action-independent control variate that removes utility already supported by the pre-rewrite memory, changing the sample-wise credit assigned to each rewrite without changing the underlying policy gradient. Unlike a learned critic, \(\Phi_t\) is obtained directly from counterfactual evaluation. A formal derivation is provided in \appref{app:value_interpretation}.

\subsection{Policy optimization objective for memory gain}

The equivalence above applies to the underlying expected policy gradient. In finite-sample optimization, however, earlier rewrites span longer remaining horizons and therefore produce returns with systematically different scales. To calibrate returns across positions, we center each return using a position-specific historical Exponential Moving Average (EMA) $b(t)$ and normalize by a global EMA scale:
\begin{equation}
    A_t = \frac{G_t-b(t)}{\sqrt{\sigma^2+\varepsilon}}.
\end{equation}  
Here, $b(t)$ tracks the historical mean return at position $t$, while $\sigma^2$ tracks a global mean squared residual. This calibration accounts for systematic horizon-dependent differences in return scale. 

We then use the position-calibrated advantage $A_t$ to optimize the memory writer. We apply a PPO-style token-level clipped policy objective \citep{schulman2017proximal}, assigning the rewrite-specific advantage $A_t$ to each token generated in $m_t$:
\begin{equation}
\begin{aligned}
J(\theta)
={}&
\mathbb{E}\Bigg[
\frac{1}{N}\sum_{t=1}^{N}\frac{1}{L_t}
\sum_{l=1}^{L_t}
\min\Big(
\rho_{t,l}(\theta)A_t, \operatorname{clip}\!\big(
\rho_{t,l}(\theta),
1-\epsilon_c,
1+\epsilon_c
\big)A_t
\Big)
\Bigg] \\
&-
\beta_{\mathrm{KL}}
\mathbb{E}\left[
\hat D_{\mathrm{KL}}
(\pi_\theta\Vert\pi_{\mathrm{ref}})
\right].
\end{aligned}
\end{equation}
Here, $\rho_{t,l}(\theta)$ is the token-level importance ratio between the current and rollout policies. The KL term regularizes the writer toward the reference policy. Length normalization by \(1/L_t\) averages token contributions within each rewrite. A detailed algorithm is provided in \appref{app:algorithm}.

\section{Experiments}
\label{sec:experiments}
\subsection{Experimental setup}
\label{sec:setup}

\textbf{Training setup.}
We instantiate MGPO through document-level IE and construct the target utility as an additive first-order surrogate of document-level micro-F1. Gold labels are assigned to the earliest observable chunk, yielding additive chunk-wise utilities aligned with the document-level objective. On held-out trajectories, the surrogate closely tracks the exact document-level F1 change (Spearman \(\rho=0.988\)). Full derivations and scoring details are provided in \autoref{app:docIE_inst}. 

\textbf{Datasets.}
We study document-level IE primarily on SciREX \citep{jain2020scirex}, which annotates \emph{Method}, \emph{Task}, \emph{Material}, and \emph{Metric} entities and their document-level structure. We optimize two tasks: \emph{salient clustering}, which groups coreferent mentions involved in reported results, and \emph{binary relation extraction} among salient entities. We train on the training split, select models on the development split, and report results on the test split.

To evaluate cross-domain transfer, we use AIPAN-10K~\citep{tang2026decoding}, a corpus of website privacy policies with annotations of mentioned data practices. Compared with SciREX, AIPAN-10K differs substantially in domain and extraction schema, and contains longer documents with denser annotations. All memory policies are trained only on SciREX and transferred to AIPAN-10K without further training or adaptation.

\textbf{Metrics and evaluation setup.}
We evaluate structured IE through entity-cluster and relation extraction, where entity clusters group mentions by entity identity in SciREX
or by privacy-data and practice categories in AIPAN-10K, and relations
capture document-supported binary associations between these clusters. This evaluation provides a consistent measure of how well each memory state supports entity and relation extraction. We additionally report the official SciREX relation metric for benchmark comparability in \appref{app:scirex-span}. We report micro-pooled precision, recall, and F1 for each extraction type. 

Documents are segmented into 1{,}024-token chunks, and the persistent memory is bounded to 256 tokens. Unless otherwise specified, all learned memory policies use Qwen3-8B and all frozen readers use Qwen3-14B \citep{yang2025qwen3}, both in non-thinking mode. Detailed scoring and implementation settings are provided in \autoref{app:experimental_details}.

\textbf{Baselines.}
We compare against three groups of baselines: (1) direct readout methods, including Direct Readout and R1-RE \cite{dai2026r1}; (2) external memory methods, including LightRAG \cite{guo2025lightrag} and Mem0 \cite{chhikara2025mem0}; and (3) learned memory policies, including MemAgent \cite{yu2026memagent} and HiMPO \citep{yan2026himpo}.

\subsection{Comparison of Memory systems}
\label{sec:main-results}

\begin{table*}[t]
\caption{
Document-level IE results on SciREX and AIPAN-10K. All memory policies are trained only on SciREX. SciREX reports in-domain performance, while AIPAN-10K evaluates out-of-domain (OOD) transfer across domain and extraction schema without further training or adaptation. Memory length is the average number of tokens in the memory state provided to the frozen reader for each chunk. Results are averaged over 4 evaluation runs. The $14B$ subscript denotes evaluation with the frozen Qwen3-14B reader in place of the method's learned reader.
}
\label{tab:main}
\centering
\small
\setlength{\tabcolsep}{3pt}
\renewcommand{\arraystretch}{0.94}

\resizebox{\textwidth}{!}{%
\begin{tabular}{l c ccc ccc c ccc ccc}
\toprule
&
\multicolumn{7}{c}{\textbf{SciREX (in-domain)}}
& \multicolumn{7}{c}{\textbf{AIPAN-10K (OOD)}} \\

\cmidrule(lr){2-8}
\cmidrule(lr){9-15}

\multirow{2}{*}[-0.6ex]{Method}
& \multirow{2}{*}[-0.6ex]{\shortstack{Memory\\length}}
& \multicolumn{3}{c}{Entity cluster}
& \multicolumn{3}{c}{Binary relation}
& \multirow{2}{*}[-0.6ex]{\shortstack{Memory\\length}}
& \multicolumn{3}{c}{Entity cluster}
& \multicolumn{3}{c}{Binary relation} \\

\cmidrule(lr){3-5}
\cmidrule(lr){6-8}
\cmidrule(lr){10-12}
\cmidrule(lr){13-15}

&
& P & R & F1
& P & R & F1
& 
& P & R & F1
& P & R & F1 \\

\midrule
\multicolumn{15}{l}{\textit{Direct Readout}} \\

Direct Readout & --
& 29.5 & 58.4 & 39.2
& 10.8 & 35.0 & 16.5
& --
& 38.0 & 31.6 & 34.5
& 24.2 & 9.4 & 13.5 \\

R1-RE & --
& 43.9 & 59.8 & 50.6
& 19.2 & 24.9 & 21.7
& --
& 28.7 & 4.6 & 7.9
& 6.0 & 0.7 & 1.2 \\

\midrule
\multicolumn{15}{l}{\textit{External Memory}} \\

LightRAG & 256
& 27.4 & 58.9 & 37.4
& 10.2 & 32.0 & 15.4
& 253
& 25.0 & 86.8 & 38.8
& 15.7 & 31.7 & 21.0 \\

Mem0 & 213
& 34.6 & 66.9 & 45.6
& 15.8 & 41.6 & 22.9
& 236
& 22.2 & 86.8 & 35.4
& 15.5 & 33.9 & 21.3 \\

\midrule
\multicolumn{15}{l}{\textit{Learned Memory}} \\

MemAgent & 62
& 49.8 & 50.1 & 49.9
& 22.6 & 16.7 & 19.2
& 200
& 24.7 & 18.9 & 21.4
& 4.4 & 4.3 & 4.4 \\

MemAgent$_{14B}$ & 64
& 50.5 & 51.3 & 50.9
& 24.1 & 25.9 & 24.9
& 202
& 31.9 & 25.2 & 28.2
& 4.5 & 1.5 & 2.2 \\

HiMPO & 256
& 39.8 & 47.6 & 43.3
& 15.4 & 17.7 & 16.5
& 256
& 21.7 & 9.9 & 13.5
& 2.1 & 1.8 & 1.9 \\

HiMPO$_{14B}$ & 256
& 35.0 & 40.8 & 37.7
& 13.5 & 16.8 & 15.0
& 256
& 33.5 & 17.8 & 23.2
& 7.1 & 0.8 & 1.4 \\

\midrule

\multicolumn{15}{l}{\textit{Decoupled Memory}} \\

MGPO-base & 202
& 27.8 & 59.2 & 37.8
& 10.0 & 34.0 & 15.5
& 241
& 24.7 & 86.6 & 38.4
& 15.5 & 33.7 & 21.3 \\

\textbf{MGPO} & \textbf{43}
& 44.2 & 61.0 & \textbf{51.3}
& 20.8 & 36.8 & \textbf{26.5}
& \textbf{168}
& 25.1 & 85.9 & \textbf{39.2}
& 17.7 & 32.9 & \textbf{22.9} \\

\bottomrule
\end{tabular}%
}
\end{table*}

\autoref{tab:main} evaluates the end-to-end performance of memory systems under their respective training settings. MemAgent and HiMPO denote results using the originally learned readers, while MemAgent$_{14B}$ and HiMPO$_{14B}$ use the same frozen Qwen3-14B reader as other baselines for a more controlled comparison. MGPO achieves the highest F1 among the compared methods. Relative to Direct Readout, MGPO improves entity cluster F1 by 30.9\% and binary relation F1 by 60.6\%. Notably, these gains coincide with a substantial reduction in the memory footprint, cutting the average memory length by 78.7\% compared with MGPO-base. Thus, MGPO improves extraction while using substantially less memory, suggesting more selective and effective retention of downstream-relevant information.

The learned memory writer also transfers to a new document domain and extraction schema. On AIPAN-10K, MGPO substantially outperforms other learned memory policies without further adaptation and exceeds the next best external-memory baseline F1 by 0.4 points on entity clusters and 1.6 points on binary relations. Repetitive outputs and format violations contribute to the low extraction scores of R1-RE, MemAgent, and HiMPO. These results suggest that the writer learns to preserve downstream-relevant information beyond the scientific-document setting on which it was trained.

\subsection{Long-horizon counterfactual credit improves memory learning}\label{sec:long_horizon_credit}
We next ask what form of credit assignment is required to learn useful memory rewrites. We compare two reference baselines with four reward variants that isolate different aspects of rewrite-level credit. \textsc{No Memory} performs chunk-wise extraction without persistent memory, while \textsc{Base Policy} uses the unoptimized writer. \textsc{Terminal} optimizes only the final document score. \textsc{Factual} uses reward-to-go over downstream factual extraction scores without subtracting utility already present before a rewrite. \textsc{Myopic} also uses reward-to-go, but measures each rewrite's marginal effect only on the current target. \textsc{Full} uses our proposed MGPO objective. For the specific implementation, please refer to \autoref{app:def_credit}.

\begin{figure}[t]
\centering
\includegraphics[width=\linewidth]{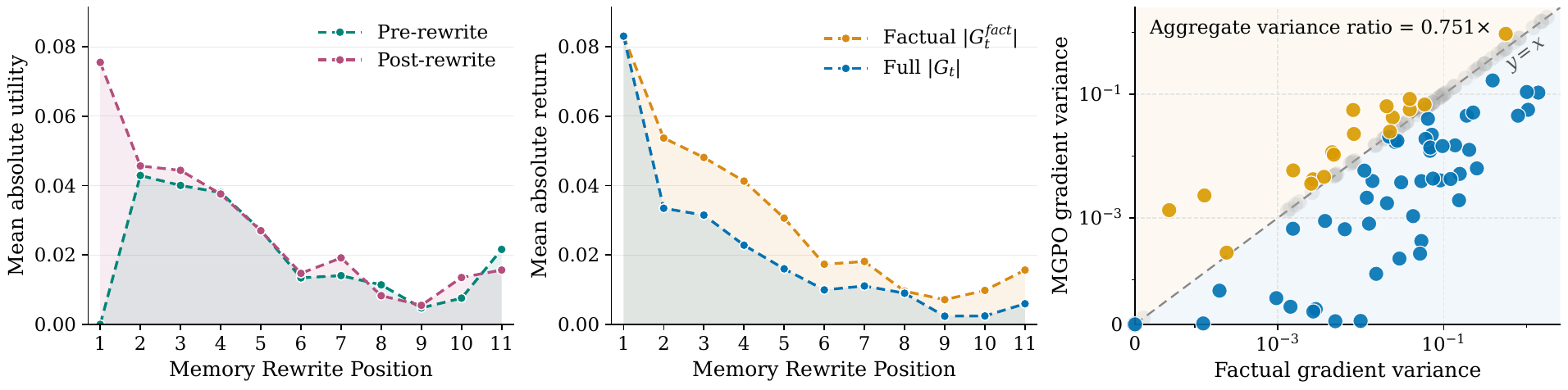}
\caption{
Inherited utility and gradient-estimator variance under the untrained memory policy on SciREX test. \textbf{Left:} Mean absolute utility before and after each memory rewrite, showing the utility already present in the preceding memory state.
\textbf{Mid:} Mean absolute factual and MGPO returns.
\textbf{Right:} Paired gradient-variance estimates at 261 fixed states from 66 documents. Variance is measured over 12,288 selected normalization parameters.}
\label{fig:inherited_utility}
\end{figure}

\paragraph{Inherited utility confounds rewrite credit.}
Before comparing the training objectives, we first examine whether the utility of an updated memory state accurately reflects the contribution of its most recent rewrite. As shown in \autoref{fig:inherited_utility}, pre- and post-rewrite utilities closely track each other once memory is established, indicating that substantial utility is inherited from the preceding state. Removing this inherited component substantially changes the return, highlighting the importance of distinguishing memory state utility from rewrite-level credit. Under a fixed, untrained memory policy, MGPO reduces aggregate across-rollout gradient variance by 24.9\% (95\% CI: 6.4\%–42.6\%), supporting the role of the pre-rewrite baseline in reducing estimator variance in this setting. Full evaluation details are provided in \appref{app:variance}.

\begin{figure} [h]
    \centering
    \includegraphics[width=0.98\linewidth]{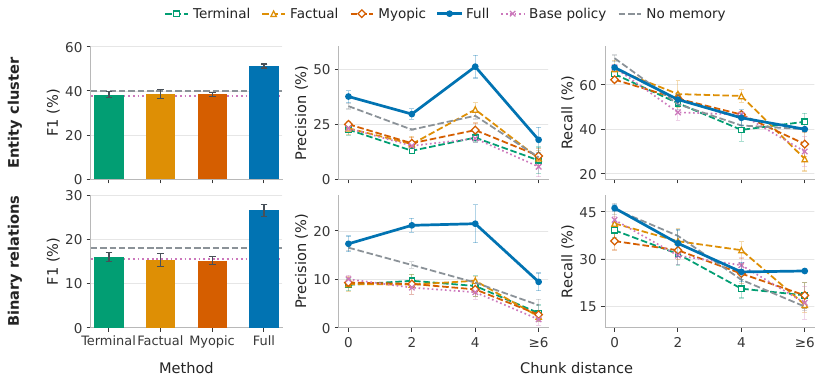}
    \caption{Effect of credit assignment on document-level extraction. \textbf{Left:} overall F1. \textbf{Middle and right:} precision and recall versus chunk distances, defined as the number of chunks separating the evidence required for a prediction. Error bars are standard deviations for 4 evaluation runs.} 
    \label{fig:credit_perf}
\end{figure}

\paragraph{Credit assignment ablations.}
The left panels of \autoref{fig:credit_perf} show that \textsc{Full} is the only variant that consistently improves over \textsc{No Memory} on both entity cluster and binary relation extraction. The key comparison is \textsc{Full} versus \textsc{Factual}: both optimize the same underlying factual objective, but \textsc{Full} subtracts utility already present before each rewrite using a pre-rewrite counterfactual baseline. The resulting performance gap supports the importance of attributing utility to the rewrite itself rather than to inherited memory content. In contrast, \textsc{Myopic} restricts marginal credit only to the current target, while \textsc{Terminal} provides only a sparse end-of-document signal. 

The distance-stratified results in the middle and right panels further show where this improvement arises.
\textsc{Full} achieves substantially higher precision across chunk distances, with the clearest separation in precision at longer distances and smaller differences in recall. This pattern suggests that long-horizon counterfactual credit improves the selectivity with which previously stored information is reused, particularly across distant chunks.

\begin{figure}[t]
  \centering
  \includegraphics[width=\linewidth]{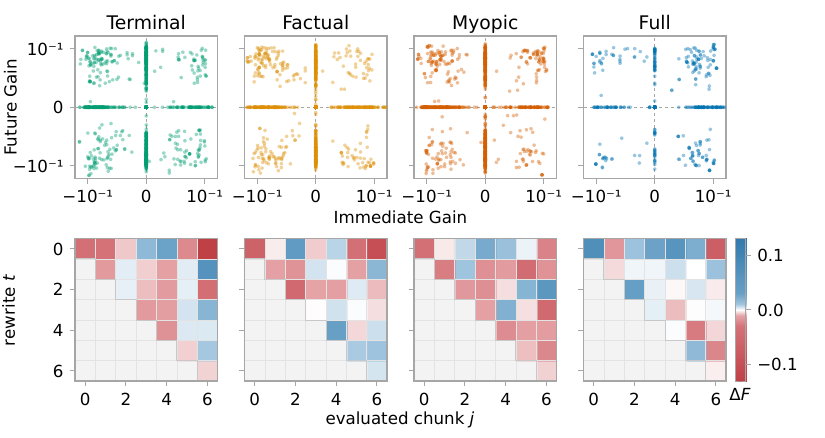}
  \caption{How credit assignment shapes memory rewrites.
\textbf{Top}: Immediate versus future extraction gain for each rewrite.
\textbf{Bottom}: Counterfactual contribution of each rewrite to subsequent chunks.}
  \label{fig:causal_mechanism}
\end{figure}

\paragraph{How credit shapes memory rewrites.}
For each learned rewrite, we compare extraction with the post-rewrite memory against the pre-rewrite memory, measuring its marginal effect on the current and subsequent targets. \autoref{fig:causal_mechanism} shows how these effects differ across objectives. In the top panel, \textsc{Factual} and \textsc{Myopic} often produce rewrites whose immediate and future utility are misaligned. \textsc{Full} shifts the distribution toward positive future utility and reduces rewrites that help the current target but hurt later ones. It also preserves rewrites with negative immediate gains but positive future rewards, consistent with learning to retain information whose value emerges through later reuse. 

The bottom panel shows these effects across subsequent chunks. Compared with the other objectives, \textsc{Full} produces stronger positive off-diagonal effects, indicating that a rewrite can continue to benefit multiple subsequent targets. These effects persist across several chunks, providing further evidence that rewrites preserve information that remains useful over time.

\subsection{MGPO Learns Reusable Memory Representations}

A key benefit of decoupling the memory writer from the downstream reader is that the learned writer can be reused across models without retraining. As shown in \autoref{tab:reader-f1}, a single MGPO-trained 8B writer, optimized with a frozen Qwen3-14B reader, improves six readers spanning four model families over their corresponding no-memory baselines. For example, entity cluster F1 improves from 34.3 to 44.5 with Qwen3-8B, from 40.0 to 51.3 with Qwen3-14B, and from 38.3 to 46.6 with Phi-4-14B. These results suggest that MGPO learns a reusable memory representation rather than a memory policy specialized to a particular reader.

This reuse is not limited to the downstream reader. As shown in \autoref{tab:main}, the same SciREX-trained writer also transfers across document domains and extraction schemas without further optimization. We further test whether this reuse extends beyond information extraction on BABILong \citep{kuratov2024babilong}, where MGPO improves over the unoptimized base writer on QA3 and several QA5 settings; results are reported in \appref{app:babilong}.

Taken together, these results show that MGPO learns a reusable memory writer rather than one tied to a specific reader or training domain. Its benefits extend across reader architectures, model scales, domains, extraction schemas, and selected downstream tasks beyond information extraction.

\begin{table}[t]
\centering
\small
\setlength{\tabcolsep}{5pt}
\newcommand{\sd}[1]{{\scriptsize$\pm$#1}}
\caption{
Pairing the learned memory writer with different readers. All rows use MGPO-trained memory writers. In the coupled variant, the writer and reader are trained jointly based on the same model. Results are mean $\pm$ standard deviation over 4 runs.
}
\label{tab:reader-f1}
\renewcommand{\arraystretch}{0.92}
\begin{tabular}{@{}llcccc@{}}
\toprule
\multirow{2}{*}{Setting}
& \multirow{2}{*}{Reader}
& \multicolumn{2}{c}{No Memory}
& \multicolumn{2}{c}{Memory} \\
\cmidrule(lr){3-4}
\cmidrule(l){5-6}
&
& Cluster & Relation & Cluster & Relation \\
\midrule

Coupled
& Self (8B)
& 44.7\sd{0.2} & 17.5\sd{0.8} & 44.9\sd{1.0} & 20.1\sd{1.4} \\

\midrule

\multirow{6}{*}{Decoupled}
& Qwen3-8B \citep{yang2025qwen3}
& 34.3\sd{0.2} & 12.8\sd{0.6} & 44.5\sd{1.0} & 20.4\sd{1.1} \\

& Qwen3-14B
& 40.0\sd{0.8} & 18.0\sd{0.6} & \textbf{51.3}\sd{0.9} & 26.5\sd{1.3} \\

& Qwen3-32B
& \textbf{45.2}\sd{0.2} & \textbf{21.5}\sd{0.5} & 50.7\sd{1.3} & \textbf{27.0}\sd{1.4} \\

& Llama3.1-8B \citep{grattafiori2024llama}
& 34.5\sd{1.4} & 12.9\sd{0.9} & 41.5\sd{0.9} & 18.3\sd{0.6} \\

& Nemo-12B \citep{mistral2024nemo}
& 39.3\sd{0.8} & 17.1\sd{0.6} & 42.0\sd{0.8} & 19.3\sd{1.1} \\

& Phi-4-14B \citep{abdin2024phi}
& 38.3\sd{0.5} & 15.2\sd{0.9} & 46.6\sd{0.4} & 23.7\sd{0.6} \\

\bottomrule
\end{tabular}
\end{table}

\section{Related Work}

\textbf{Learning and optimizing memory.}
Earlier approaches learn recurrent or retrieved memory representations \citep{bulatov2022recurrent, wang2023augmenting}, while EMR uses reinforcement learning to manage bounded streaming memory \citep{han2019episodic}. RECOMP and PRCA train compressors or contextual adapters for fixed downstream models \citep{xu2024recomp, yang2023prca}, and CompAct learns iterative textual compression \citep{yoon2024compact}. Previous methods optimize persistent memory through downstream rewards. MEM1 jointly learns reasoning and memory consolidation \citep{zhou2026mem1}, while Memory-R1, Mem-$\alpha$, MemAgent, and MemPO optimize memory policies through reinforcement learning \citep{yan2026memory, wang2025mem, yu2026memagent, li2026mempo}. More recent methods introduce memory-specific credit signals, including hindsight-informed utility in HiMPO \citep{yan2026himpo}, fine-grained feedback in Fine-Mem \citep{ma2026fine}, belief-entropy supervision in MMPO \citep{liu2026meta}, and shared-state local rerollouts in Memory-R2 \citep{yan2026memoryr2}. MGPO instead isolates the contribution of each rewrite from utility inherited from prior memory.

\textbf{Credit assignment for RL.}
Dense credit assignment has been studied through process rewards for intermediate reasoning steps \citep{lightman2024let, wang2024math} and progress-based rewards that measure improvement toward eventual task success \citep{setlur2025rewarding}. Long-range temporal credit has also been addressed through value transport, which propagates delayed rewards back to earlier decisions \citep{hung2019optimizing}. Counterfactual credit assignment instead  isolates an action's contribution by comparing its outcome against an appropriate counterfactual baseline, as in difference rewards \citep{tumer2002learning} and COMA \citep{foerster2018counterfactual}. This principle has been further developed for temporal credit assignment in single-agent reinforcement learning \citep{mesnard2021counterfactual}. MGPO extends counterfactual attribution to persistent memory, crediting each rewrite for the marginal utility it contributes across the remaining horizon.

\textbf{Document-level reasoning and information extraction.}
Document-level reasoning and information extraction require integrating distributed evidence through document graphs, localized aggregation, and reasoning-based architectures \citep{christopoulou2019connecting,nan2020reasoning,zhou2021document,ma2023dreeam,dai2026r1}. Recent approaches address this challenge through recurrent representations \citep{wang2023augmenting}, memory tokens \citep{gao2024ttm}, retrieval and graph/vector stores \citep{guo2025lightrag}, or explicit textual memory \citep{zhu2025llmlink}. MGPO instead learns a bounded textual memory policy for deciding what to remember as context unfolds.

\section{Conclusion}
In this paper, we introduced Memory Gain Policy Optimization (MGPO) for learning persistent memory for long-horizon tasks. MGPO assigns each memory rewrite counterfactual credit according to its marginal contribution to current and future targets, providing dense supervision for optimizing what information should persist. Our experiments show that this long-horizon marginal credit produces more selective memory, improves long-range prediction, and supports transfer across readers, domains, and downstream tasks. These results suggest that learning what to remember requires crediting each memory rewrite for the downstream utility it uniquely contributes over time.

\newpage
\bibliography{MGPO}
\bibliographystyle{iclr2027_conference}

\newpage
\appendix
\startcontents[appendix]

\section*{Appendix Contents}
\setcounter{tocdepth}{2}
\printcontents[appendix]{}{1}{}
\clearpage

\section{Theoretical Foundations of MGPO}
\label{app:value_interpretation}
\subsection{POMDP formulation.}

We formalize memory writing as an undiscounted finite-horizon contextual partially observable Markov decision process (POMDP). At the beginning of each episode, a task and its supervision are sampled as the environment context. Let $x_t$ denote the augmented pre-action state containing the task $\mathcal{T}$, stream position, pre-rewrite memory $m_{t-1}$, and relevant history, while the writer observes only $o_t=(\mathcal{T},m_{t-1},c_t)$. The action is the rewritten memory itself, $a_t=m_t\sim\pi_\theta(\cdot\mid o_t)$, after which the process advances to $x_{t+1}$ with $m_t$ as the persistent state.

Let \(F(m,c_j)\) denote the expected utility of the fixed reader on target
\(c_j\) when conditioned on memory \(m\), and write
\(F_{t,j}=F(m_t,c_j)\).
The factual reward and remaining-horizon return are
\begin{equation}
r_t^{\mathrm{fact}}=F_{t,t},
\qquad
G_t^{\mathrm{fact}}
=
\sum_{j=t}^{N}F_{j,j}.
\end{equation}
The corresponding value, action-value, and advantage functions are
\begin{align}
V_t^{\pi,\mathrm{fact}}(x)
&=
\mathbb{E}_\pi\!\left[
G_t^{\mathrm{fact}}
\mid x_t=x
\right],\\
Q_t^{\pi,\mathrm{fact}}(x,a)
&=
\mathbb{E}_\pi\!\left[
G_t^{\mathrm{fact}}
\mid x_t=x,\ a_t=a
\right],\\
A_t^{\pi,\mathrm{fact}}(x,a)
&=
Q_t^{\pi,\mathrm{fact}}(x,a)-V_t^{\pi,\mathrm{fact}}(x).
\end{align}

\subsection{Memory Gain as Potential-Based Reward Shaping}
For the pre-action state \(x_t\) with memory \(m_{t-1}\), define
\begin{equation}
\Phi_t(x_t)
=
\sum_{j=t}^{N}F_{t-1,j},
\qquad
\Phi_{N+1}=0.
\end{equation}
Thus, \(\Phi_t(x_t)\) is the remaining-horizon utility of the fixed
pre-rewrite memory. Unlike \(V_t^\pi\), it does not average over future
policy-induced memory updates, but is directly evaluated from the
counterfactual matrix. Since it is determined by \(x_t\), it is fixed
before \(a_t\) is sampled.

\paragraph{Lemma 1 (Potential form of Memory Gain).}
Memory Gain is a potential-shaped version of the factual streaming reward:

\begin{equation}
MG_t=r_t^{\mathrm{fact}}+\Phi_{t+1}(x_{t+1})-\Phi_t(x_t).
\end{equation}

\textit{Proof.}
Since $x_{t+1}$ contains the updated memory $m_t$, we have $\Phi_{t+1}(x_{t+1})=\sum_{j=t+1}^{N}F_{t,j}$. Therefore,

\begin{align}
    r_t^{\mathrm{fact}}+\Phi_{t+1}(x_{t+1})-\Phi_t(x_t)&=F_{t,t}+\sum_{j=t+1}^{N}F_{t,j}-\sum_{j=t}^{N}F_{t-1,j}\\
    &=\sum_{j=t}^{N}\left(F_{t,j}-F_{t-1,j}\right) \\
    &=MG_t.
\end{align}

Thus, in the undiscounted finite-horizon setting, Memory Gain has the standard form of potential-based reward shaping \citep{ng1999policy}. Because the stream position is included in the augmented state $x_t$, the time-dependent quantity $\Phi_t$ can be treated as an ordinary state potential on this augmented state space. \hfill$\square$
\subsection{Return Decomposition and Advantage Preservation}

\paragraph{Proposition 1 (Return decomposition and advantage preservation).}
For the Memory-Gain return
\begin{equation}
G_t^{\mathrm{MG}}=\sum_{k=t}^{N}MG_k,
\end{equation}
we have
\begin{equation}
G_t^{\mathrm{MG}}
=
G_t^{\mathrm{fact}}-\Phi_t(x_t),
\end{equation}
and consequently
\begin{equation}
A_t^{\pi,\mathrm{MG}}(x,a)
=
A_t^\pi(x,a).
\end{equation}

\textit{Proof.}
By Lemma~1,
\begin{align}
G_t^{\mathrm{MG}}
&=
\sum_{k=t}^{N}
\left(
r_k^{\mathrm{fact}}
+\Phi_{k+1}(x_{k+1})
-\Phi_k(x_k)
\right) \\
&=
\sum_{k=t}^{N}r_k^{\mathrm{fact}}
-\Phi_t(x_t)
+\Phi_{N+1}(x_{N+1}) \\
&=
G_t^{\mathrm{fact}}-\Phi_t(x_t),
\end{align}
where $\Phi_{N+1}=0$.

Taking conditional expectations yields
\begin{align}
Q_t^{\pi,\mathrm{MG}}(x,a)
&=
\mathbb{E}_\pi
\left[
G_t^{\mathrm{MG}}
\mid x_t=x,a_t=a
\right] \\
&=
\mathbb{E}_\pi
\left[
G_t^{\mathrm{fact}}-\Phi_t(x_t)
\mid x_t=x,a_t=a
\right] \\
&=
Q_t^\pi(x,a)-\Phi_t(x),
\end{align}
and similarly,
\begin{align}
V_t^{\pi,\mathrm{MG}}(x)
&=
\mathbb{E}_\pi
\left[
G_t^{\mathrm{MG}}
\mid x_t=x
\right] \\
&=
V_t^\pi(x)-\Phi_t(x).
\end{align}
Therefore,
\begin{align}
A_t^{\pi,\mathrm{MG}}(x,a)
&=
Q_t^{\pi,\mathrm{MG}}(x,a)
-
V_t^{\pi,\mathrm{MG}}(x) \\
&=
Q_t^\pi(x,a)-V_t^\pi(x) \\
&=
A_t^\pi(x,a).
\end{align}
\hfill$\square$

\subsection{Policy-Gradient Preservation}
\paragraph{Proposition 2 (Policy-gradient preservation).}
We define the action log-probability gradient:
\begin{equation}
\psi_t
=
\nabla_\theta
\log\pi_\theta(a_t\mid o_t).
\end{equation}
Then the Memory-Gain return induces the same expected score-function
policy gradient as the factual return:
\begin{equation}
\nabla_\theta J_{\mathrm{MG}}=
\mathbb{E}_\pi
\left[
\sum_{t=1}^{N}
\psi_t G_t^{\mathrm{MG}}
\right]
=
\mathbb{E}_\pi
\left[
\sum_{t=1}^{N}
\psi_t G_t^{\mathrm{fact}}
\right]
=
\nabla_\theta J_{\mathrm{fact}}.
\end{equation}

\textit{Proof.}
From Proposition~1,
\begin{equation}
G_t^{\mathrm{MG}}
=
G_t^{\mathrm{fact}}-\Phi_t(x_t).
\end{equation}
Hence,
\begin{align}
\mathbb{E}_\pi
\left[
\sum_{t=1}^{N}
\psi_t G_t^{\mathrm{MG}}
\right]
&=
\mathbb{E}_\pi
\left[
\sum_{t=1}^{N}
\psi_t
\left(
G_t^{\mathrm{fact}}-\Phi_t(x_t)
\right)
\right] \\
&=
\mathbb{E}_\pi
\left[
\sum_{t=1}^{N}
\psi_t G_t^{\mathrm{fact}}
\right]
-
\sum_{t=1}^{N}
\mathbb{E}_\pi
\left[
\psi_t\Phi_t(x_t)
\right].
\end{align}

By the law of iterated expectation,
\begin{align}
\mathbb{E}_\pi
\left[
\psi_t\Phi_t(x_t)
\right]
&=
\mathbb{E}_\pi
\left[
\mathbb{E}_\pi
\left[
\psi_t\Phi_t(x_t)
\mid x_t
\right]
\right].
\end{align}
Since $\Phi_t(x_t)$ is determined by the pre-action state,
\begin{align}
\mathbb{E}_\pi
\left[
\psi_t\Phi_t(x_t)
\mid x_t
\right]
&=
\Phi_t(x_t)
\mathbb{E}_\pi
\left[
\psi_t
\mid x_t
\right] \\
&=
\Phi_t(x_t)
\mathbb{E}_{a_t\sim\pi_\theta(\cdot\mid o_t)}
\left[
\nabla_\theta
\log\pi_\theta(a_t\mid o_t)
\right].
\end{align}
Because $o_t$ is determined by $x_t$,
\begin{align}
\mathbb{E}_{a_t\sim\pi_\theta(\cdot\mid o_t)}
\left[
\nabla_\theta
\log\pi_\theta(a_t\mid o_t)
\right]
&=
\sum_a
\pi_\theta(a\mid o_t)
\nabla_\theta
\log\pi_\theta(a\mid o_t) \\
&=
\sum_a
\nabla_\theta
\pi_\theta(a\mid o_t) \\
&=
\nabla_\theta
\sum_a
\pi_\theta(a\mid o_t) \\
&=
\nabla_\theta 1 \\
&=
0.
\end{align}
Therefore,
\begin{equation}
\mathbb{E}_\pi
\left[
\psi_t\Phi_t(x_t)
\right]
=
0
\end{equation}
for every $t$, and thus
\begin{equation}
\mathbb{E}_\pi
\left[
\sum_{t=1}^{N}
\psi_tG_t^{\mathrm{MG}}
\right]
=
\mathbb{E}_\pi
\left[
\sum_{t=1}^{N}
\psi_tG_t^{\mathrm{fact}}
\right].
\end{equation}
The right-hand side is the standard score-function gradient of
$J_{\mathrm{fact}}$, which proves the result.
\hfill$\square$

\paragraph{Remark (Episode-level objective equivalence).}
Telescoping over the complete trajectory gives
\begin{equation}
G_1^{\mathrm{MG}}
=
G_1^{\mathrm{fact}}-\Phi_1(x_1),
\end{equation}
and hence
\begin{equation}
J_{\mathrm{MG}}(\theta)
=
J_{\mathrm{fact}}(\theta)
-
\mathbb{E}\!\left[\Phi_1(x_1)\right].
\end{equation}
Since the initial memory and episode context are sampled independently of
\(\theta\), the last term is constant with respect to \(\theta\). Therefore,
\begin{equation}
\nabla_\theta J_{\mathrm{MG}}
=
\nabla_\theta J_{\mathrm{fact}}.
\end{equation}

\paragraph{Finite-sample effect of the counterfactual potential.}
Although the expected policy gradient is preserved, the samplewise
contributions differ:
\begin{equation}
\psi_t
\left(
G_t^{\mathrm{MG}}-G_t^{\mathrm{fact}}
\right)
=
-\psi_t\Phi_t(x_t),
\qquad
\mathbb{E}\!\left[\psi_t\Phi_t(x_t)\right]=0.
\end{equation}
Thus, $\Phi_t$ acts as a counterfactual control variate: it modifies finite-sample updates while preserving their expectation, and can reduce variance without requiring a learned critic.

The position EMA provides action-independent centering, while the global EMA scale is used only as an optimization normalization. Thus, the theoretical equivalence applies to the underlying objective rather than to the exact finite-sample PPO update. Accordingly, MGPO targets the same factual streaming objective through a counterfactually centered estimator, rather than optimizing a different utility objective.

\section{MGPO Algorithm and Task Instantiation}
\label{app:docIE_inst}
\subsection{MGPO Algorithm}
\label{app:algorithm}

\begin{algorithm}[H]
\caption{Memory Gain Policy Optimization (MGPO)}
\label{alg:mgpo}
\begin{algorithmic}[1]
\Require Policy $\pi_\theta$, reference policy $\pi_{\mathrm{ref}}$,
frozen reader $\mathcal{E}$, dataset $\mathcal{D}$,
utility function $\mathcal{U}$, EMA decay $\alpha$,
KL coefficient $\beta_{\mathrm{KL}}$, learning rate $\eta$,
clipping threshold $\epsilon_c$, numerical stabilizer $\varepsilon$
\State Initialize position-wise EMA means $\{b(t)\}$ and
global EMA variance $\sigma^2$

\While{not converged}
    \State Sample a minibatch $\mathcal{B}\sim\mathcal{D}$

    \ForAll{$d\in\mathcal{B}$}
        \State Split $d$ into a stream inputs $(c_1,\ldots,c_N)$ and gold answers $(y_1,\ldots,y_N)$
        \State $m_0\gets\varnothing$

        \For{$t=1$ to $N$}
            \State $m_t\sim
            \pi_{\theta_{\mathrm{old}}}
            (\cdot\mid m_{t-1},c_t)$
            \Comment{Streaming memory rollout}
        \EndFor

        \For{$0\leq s\leq j\leq N$}
           \State $F_{s,j}\gets \mathbb{E}\left[\mathcal{U}(\mathcal{E}(m_s,c_j),y_j)\right]$
            \Comment{Counterfactual evaluation}
        \EndFor

        \For{$t=1$ to $N$}
            \State
            $MG_t\gets
            \displaystyle\sum_{j=t}^{N}
            \left(F_{t,j}-F_{t-1,j}\right)$
            \Comment{Long-horizon marginal utility}

            \State
            $G_t\gets
            \displaystyle\sum_{k=t}^{N}MG_k$

            \State
            $\delta_t\gets G_t-b(t)$
        \EndFor
    \EndFor

    \State
    $\sigma^2\gets
    \alpha\sigma^2+
    (1-\alpha)
    \underset{(d,t)\in\mathcal{B}}{\operatorname{mean}}
    \delta_{d,t}^{\,2}$
    \Comment{One global scale update per minibatch}

    \State
    $A_{d,t}\gets
    \dfrac{\delta_{d,t}}{\sqrt{\sigma^2}+\varepsilon}
    \quad\forall(d,t)\in\mathcal{B}$

    \ForAll{memory positions $t$ represented in $\mathcal{B}$}
        \State
        $b(t)\gets
        \alpha b(t)+(1-\alpha)
        \underset{d:(d,t)\in\mathcal{B}}{\operatorname{mean}}G_{d,t}$
        \Comment{One position-wise mean update per minibatch}
    \EndFor

    \State
    $\rho_{d,t,\ell}\gets
    \dfrac{
    \pi_\theta(m_{d,t,\ell}\mid
    m_{d,t-1},c_{d,t},m_{d,t,<\ell})
    }{
    \pi_{\theta_{\mathrm{old}}}(m_{d,t,\ell}\mid
    m_{d,t-1},c_{d,t},m_{d,t,<\ell})
    }$

    \State
    $\mathcal{J}_{\mathrm{clip}}\gets
    \displaystyle
    \frac{1}{|\mathcal{B}|}
    \sum_{d\in\mathcal{B}}
    \frac{1}{N_d}
    \sum_{t=1}^{N_d}
    \frac{1}{L_{d,t}}
    \sum_{\ell=1}^{L_{d,t}}
    \min\!\left(
    \rho_{d,t,\ell}A_{d,t},
    \operatorname{clip}
    (\rho_{d,t,\ell},1-\epsilon,1+\epsilon)A_{d,t}
    \right)$

    \State
    $\theta\gets\theta+\eta\nabla_\theta
    \left[
    \mathcal{J}_{\mathrm{clip}}
    -\beta\,\operatorname{KL}
    (\pi_\theta\Vert\pi_{\mathrm{ref}})
    \right]$

\EndWhile
\end{algorithmic}
\end{algorithm}

\subsection{Streaming Target Construction}
We convert document-level supervision into chunk-level targets by assigning each gold structure to the earliest chunk at which it becomes observable. During training, entity-level labels are assigned to the chunk containing their first mention, while relation labels are assigned to the first chunk by which all required arguments have appeared. For evaluation, we merge the outputs from different chunks and compare them against the document-level IE gold labels. Let \(y_j^{(a)}\) denote the gold structures of task \(a\) assigned to chunk \(c_j\). We further define the historical labels at chunk \(j\) as

$$ H_j^{(a)}=\bigcup_{k<j} y_k^{(a)}. $$

When scoring chunk \(c_j\), predicted structures matching \(H_j^{(a)}\) are masked, and the remaining predictions are evaluated against \(y_j^{(a)}\). This produces a unique temporal attribution of each gold structure and avoids repeatedly scoring targets assigned to earlier chunks.

\subsection{Additive Utility for Document-Level Micro-F1}
Because the reader is frozen, the role of memory is to improve the extraction quality achievable by a fixed readout. We therefore define memory utility relative to the performance of the same reader without memory. For task \(a\), memory state \(m_t\), and target chunk \(c_j\), we summarize the resulting extraction outcome by the count vector
$$ \mathbf q_{t,j}^{(a)} = \left( TP_{t,j}^{(a)}, FP_{t,j}^{(a)}, FN_{t,j}^{(a)} \right), $$

computed from reader \(\mathcal{E}(m_t,c_j;\xi)\) under the scoring rule above. Because document-level micro-F1 is defined over counts aggregated across chunks and is therefore non-additive, we construct an additive first-order surrogate around a memory-off reference. 
Let 
$$ \mathbf Q_{\mathrm{off}}^{(a)} = \sum_{j=1}^{N} \mathbf q_{\mathrm{off},j}^{(a)} $$
denote the document-level count vector obtained with memory disabled, and define
$$ \mathbf w^{(a)} = \nabla f\!\left(\mathbf Q_{\mathrm{off}}^{(a)}\right), \qquad f(TP,FP,FN) = \frac{2TP}{2TP+FP+FN}. $$

The utility assigned to counterfactual cell \((t,j)\) is then
$$ F_{t,j} = \sum_a \lambda_a \left\langle \mathbf w^{(a)}, \mathbf q_{t,j}^{(a)} \right\rangle, $$

where \(\lambda_a\) is the weight of task \(a\). For training, we set $\lambda^{\text{entity}}=\lambda^{\text{relation}}=0.5$. Since \(\mathbf w^{(a)}\) is fixed for a given document and task, these cell utilities are additive across target chunks. For each document, the memory-off reference and its linearization weights are fixed across all sampled memory states, allowing their utilities to be compared under the same reader and document-level objective. The relation between this surrogate and exact document-level micro-F1 is derived below.

\subsection{Relation to Document-Level Micro-F1}
\label{app:linearized_score}

We provide the derivation of the linearized score used in \autoref{sec:credit_assign} and establish its relation to document-level F1 improvement.

For a single target type, define
\begin{equation}
f(TP,FP,FN)=\frac{2TP}{D}, \qquad D=2TP+FP+FN.
\label{eq:app_f1}
\end{equation}
Its partial derivatives are
\begin{equation}
\frac{\partial f}{\partial TP}=\frac{2(1-f)}{D}, \qquad \frac{\partial f}{\partial FP}=-\frac{f}{D}, \qquad \frac{\partial f}{\partial FN}=-\frac{f}{D}.
\label{eq:app_f1_gradient}
\end{equation}
Accordingly, for an offline anchor $\mathbf Q_{\mathrm{off}}=(TP_0,FP_0,FN_0)$ with $D_0=2TP_0+FP_0+FN_0$ and $f_0=2TP_0/D_0$, the fixed linearization weights are
\begin{equation}
\mathbf w=\left(\frac{2(1-f_0)}{D_0},-\frac{f_0}{D_0},-\frac{f_0}{D_0}\right).
\label{eq:app_anchor_weights}
\end{equation}
Thus, the local approximation rewards additional true positives while penalizing false positives and false negatives according to the behaviors of the frozen reader.

A first-order Taylor expansion around $\mathbf Q_{\mathrm{off}}$ gives
\begin{equation}
f(\mathbf Q)=f(\mathbf Q_{\mathrm{off}})+\mathbf w^\top\left(\mathbf Q-\mathbf Q_{\mathrm{off}}\right)+O\!\left(\left\|\mathbf Q-\mathbf Q_{\mathrm{off}}\right\|^2\right).
\label{eq:app_taylor}
\end{equation}
This motivates the relative cell contribution
\begin{equation}
\widetilde F_{t,j}=\mathbf w^\top\left(\mathbf q_{t,j}-\mathbf q_{0,j}\right).
\label{eq:app_relative_score}
\end{equation}
The explicit subtraction of the offline contribution is unnecessary for counterfactual differences because it cancels between adjacent memory states:
\begin{equation}
\widetilde F_{t,j}-\widetilde F_{t-1,j}=\mathbf w^\top\left(\mathbf q_{t,j}-\mathbf q_{t-1,j}\right)=F_{t,j}-F_{t-1,j}.
\label{eq:app_baseline_cancel}
\end{equation}
Hence, the main text uses the simpler absolute form $F_{t,j}=\mathbf w^\top\mathbf q_{t,j}$ without altering the resulting memory gain.

Finally, the accumulated memory gain admits a telescoping interpretation. Let $\mathbf Q_{\mathrm{pipe}}^{(a)}$ denote the document-level count vector produced along the deployment trajectory for task $a$. Then
\begin{equation}
\sum_{t=1}^{N}MG_t=\sum_{j=1}^{N}\left(F_{j,j}-F_{0,j}\right)=\sum_a\lambda_a\left\langle\mathbf w^{(a)},\mathbf Q_{\mathrm{pipe}}^{(a)}-\mathbf Q_{\mathrm{off}}^{(a)}\right\rangle.
\label{eq:app_mg_telescoping}
\end{equation}
Applying the same first-order approximation to each target type yields
\begin{equation}
\sum_{t=1}^{N}MG_t\approx\sum_a\lambda_a\left[f\!\left(\mathbf Q_{\mathrm{pipe}}^{(a)}\right)-f\!\left(\mathbf Q_{\mathrm{off}}^{(a)}\right)\right].
\label{eq:app_mg_f1}
\end{equation}
Therefore, $F_{t,j}$ provides an additive local surrogate for document-level extraction quality, while $MG_t$ attributes changes in that surrogate to individual memory rewrites. Their accumulated credit approximates the improvement of the streaming pipeline over the memory-off reader under the original document-level F1 objective.

\section{Experimental Setup and Reproducibility}
\label{app:experimental_details}

\subsection{Datasets and Splits}
We evaluate on SciREX test set and a 50-document subset of AIPAN-10K selected from 9,430 annotated policies after length, annotation-consistency, relation-density, and deduplication filtering. Dataset statistics are reported in (\autoref{tab:dataset_statistics}). For AIPAN-10K, gold mention boundaries are provided to isolate document-level clustering and relation extraction. This focuses evaluation on document-level entity clustering and relation extraction, since our objective does not optimize mention identification. 

\begin{table*}[h]
\centering
\small
\caption{IE Test-set statistics. Entity and relation counts are per-document averages over entity clusters and unique binary relations.}
\label{tab:dataset_statistics}
\begin{tabular}{lrrrrrrr}
\toprule
Dataset & \# Docs & Avg. length & Max. length  & Avg. entities & Avg. relations \\
\midrule
SciREX    & 66 & 6,947  & 16,908 & 5.12  & 8.67  \\
AIPAN-10K & 50 & 15,375 & 59,344  & 28.00 & 65.54 \\
\bottomrule
\end{tabular}
\end{table*}

We evaluate BABILong QA1--QA5 at 8K, 16K, 32K, and 128K contexts using 100 examples per task and length, reporting mean accuracy under the official scoring procedure. In order to test only memory capabilities, all methods use their trained memory writer, while sharing the same frozen Qwen3-14B reader.

\subsection{Implementation Protocols}
\label{app:protocol}

\paragraph{Baseline implementation.}
\autoref{tab:baseline_protocol} summarizes the deployment settings. Direct Readout and R1-RE process the entire document in a single pass. LightRAG and Mem0 construct external memory stores and provide retrieved content to a frozen reader, with the memory budget applied to the retrieved text. MemAgent and HiMPO update memory after each chunk and use either their own trained model or the shared frozen Qwen3-14B reader for readout. MGPO updates memory sequentially and uses the frozen Qwen3-14B reader for downstream predictions. We use 1,024-token chunks with a 256-token memory budget for SciREX and AIPAN-10K, and 4,096-token chunks with a 1,024-token memory budget for BABILong.

\paragraph{Credit assignment variants.}
\label{app:def_credit}
Table~\ref{tab:credit_variants} defines the four credit-assignment variants. \textsc{Factual}, \textsc{Myopic}, and \textsc{Full} share the same chunk-aligned supervision, position-EMA calibration, and differ only in reward definition.
\begin{table}[h]
\centering
\small
\caption{Credit-assignment variants.}
\label{tab:credit_variants}
\setlength{\tabcolsep}{5pt}
\renewcommand{\arraystretch}{1.02}
\begin{tabular}{lcc}
\toprule
Method & Step reward $r_t$ & Return $G_t$ \\
\midrule
Terminal & --
         & $\mathrm{F1}_{\mathrm{doc}}$ \\
Factual  & $F_{t,t}$
         & $\sum_{k=t}^{N}F_{k,k}$ \\
Myopic   & $\Delta_{t,t}$
         & $\sum_{k=t}^{N}\Delta_{k,k}$ \\
Full     & $MG_t$
         & $\sum_{k=t}^{N}MG_k$ \\
\bottomrule
\end{tabular}
\end{table}
\paragraph{Training supervision.}
MGPO uses chunk-aligned targets deterministically derived from existing SciREX annotations, whereas \textsc{Terminal}, MemAgent and HiMPO use document-level targets. Thus, \textsc{Terminal} is used to verify the effectiveness of the chunk-aligned reward, and the controlled credit-assignment comparison is provided by \textsc{Factual}, \textsc{Myopic}, and \textsc{Full}, which share identical supervision and readout protocols.

\paragraph{Optimizer choice.} We train R1-RE and HiMPO with GRPO ($G=8$) \citep{shao2024deepseekmath}. For MemAgent, originally trained with Multi-Conv DAPO~\citep{yu2026memagent}, we retain its memory mechanism and outcome reward and use the same GRPO objective to standardize optimization across these baselines. We also evaluate a Multi-Conv DAPO variant under the same data, reward, model, and training budget. On the SciREX test set, MemAgent-GRPO achieved 49.9 salient-cluster F1 and 19.2 binary-relation F1, compared with 31.7 and 10.7 for this variant. We therefore use GRPO for MemAgent in our comparison.

\begin{table}[h]
\centering
\small
\caption{Implementation settings of baselines.}
\label{tab:baseline_protocol}
\setlength{\tabcolsep}{5pt}
\begin{tabular}{llcccc}
\toprule
\textbf{Type} &
\textbf{Method} &
\textbf{Trainable component} &
\textbf{Readout} &
\textbf{Chunk} &
\textbf{Memory budget} \\
\midrule

\multirow{2}{*}{Direct}
& Direct-Readout
& --
& Qwen3-14B
& Whole doc.
& None \\

& R1-RE
& Qwen3-8B readout
& Self
& Whole doc.
& None \\

\midrule

\multirow{2}{*}{External}
& LightRAG
& --
& Qwen3-14B
& 1,024 / 4,096
& 256 / 1,024 \\

& Mem0
& --
& Qwen3-14B
& 1,024 / 4,096
& 256 / 1,024 \\

\midrule

\multirow{2}{*}{Learned}
& MemAgent
& Qwen3-8B writer
& Self / Qwen3-14B
& 1,024 / 4,096
& 256 / 1,024 \\

& HiMPO
& Qwen3-8B writer
& Self / Qwen3-14B
& 1,024 / 4,096
& 256 / 1,024 \\

\midrule

Ours
& MGPO
& Qwen3-8B writer
& Qwen3-14B
& 1,024 / 4,096
& 256 / 1,024 \\

\bottomrule
\end{tabular}
\end{table}

\subsubsection{Evaluation protocol.}
We evaluate entity clusters and binary relations using dataset-specific cluster matching (Table~\ref{tab:scoring}). Relations are scored through independently matched relation-referenced entities. We report micro-precision, recall, and F1 from pooled TP/FP/FN counts. Incomplete or incorrect JSON formatting is treated as a format violation, and the output is recorded as null.

\begin{table}[t]
\centering
\small
\setlength{\tabcolsep}{4pt}
\renewcommand{\arraystretch}{1.1}
\begin{tabularx}{\linewidth}{@{}p{0.19\linewidth}X@{}}
\toprule
\textbf{Dataset} & \textbf{Cluster matching} \\
\midrule
SciREX &
Greedy one-to-one matching by gold-span coverage ($\geq 0.5$),
without an entity-type constraint. \\
\addlinespace[3pt]
AIPAN-10K &
Same-type matching using exact normalized aliases for data types
and token-F1 ($\geq 0.5$) for other entity types. Many-to-one
matches are allowed. \\
\bottomrule
\end{tabularx}
\caption{Dataset-specific cluster matching for document-level evaluation.}
\label{tab:scoring}
\end{table}

\textbf{SciREX.} We evaluate salient-cluster identification and binary relation extraction. Gold-span coverage is the fraction of annotated spans in a gold cluster overlapped by at least one predicted span. Binary relations are untyped and are induced from pairs of salient entities that co-occur in gold result tuples. We do not evaluate 4-ary tuple extraction, which additionally requires composing pairwise relations into globally consistent result structures. 

\textbf{AIPAN-10K.}
We provide gold mention boundaries to isolate document-level clustering and relation extraction. Because its clusters represent semantic categories rather than coreference identities, multiple predicted subclusters may map to the same gold category. We therefore allow many-to-one matching. Data types use normalized exact matching, while other entity types use token-F1 $\geq0.5$ for long phrases matching. Mapped relation pairs are deduplicated before scoring.

\textbf{BABILong.}
We evaluate question-answering accuracy using the benchmark's official answer-matching procedure. Answers are generated from the final memory and the question, without access to the original document.

\subsection{Hyperparameter Settings.}
\label{app:hyperparameters}
We assign equal weights to entity and relation utility, $\lambda^{\mathrm{entity}}=\lambda^{\mathrm{relation}}=0.5$,
and use undiscounted returns ($\gamma=1$). For position-calibrated advantages, we set the EMA decay to $\alpha=0.9$ and the numerical stabilizer to $\varepsilon=10^{-6}$. All zero-based positions $t\geq9$ share one baseline bucket.

We optimize the writer with AdamW, using a peak learning rate of $10^{-6}$, 10 warmup steps, and cosine decay to $10^{-7}$. The optimizer uses betas $(0.9,0.999)$, weight decay $0.01$, and gradient-norm clipping at $1.0$.
We set the PPO clipping threshold to $\epsilon=0.2$ and the KL coefficient to $\beta=10^{-3}$. The KL penalty uses the low-variance estimator and is added to the loss; no entropy bonus is used.

We train for one epoch, with 4 documents per rollout batch and one sampled trajectory per document. Each batch receives one PPO epoch, with minibatches of two memory-update sequences. Writer rollouts use temperature $1.0$ and top-$p=1.0$ during  training, and temperature $0.7$, top-$p=0.8$, and top-$k=20$ during evaluation. The Qwen3-14B reader uses temperature $0.7$, top-$p=0.8$, and top-$k=20$ for both training utility estimation and test readout. Others use the official recommended settings.

\subsection{Additional Experiment Details.}
\label{app:variance}
\paragraph{Inherited utility.}
We evaluate the untrained Qwen3-8B writer using four trajectories for each of 63 SciREX test documents with complete records. Pre- and post-rewrite utilities are evaluated by holding the respective memory fixed over remaining targets, using one reader sample with seed~0. At each position, we average absolute utilities first within documents and then equally across documents containing that position. Using the same trajectories and averaging procedure, we compare mean absolute factual and MGPO suffix returns. Counterfactual baseline subtraction precedes the absolute value.

\paragraph{Gradient variance.}
On the same SciREX test set, we select four approximately evenly spaced positions, including the first and last, from one trajectory per test document. At each fixed state $(m_{t-1},c_t)$, we sample four continuations using the training decoding settings and retain states with at least two valid branches, yielding 261 states. Both estimators share the same branches and current-action score gradients. Utility evaluations average eight reader samples with matched seeds~1--8 and cache identical requests, fixing the pre-rewrite baseline across branches. The first position instead reuses the seed-0 empty-memory anchor. Variance is the trace of the unbiased sample covariance over 12,288 parameters comprising the final block's two normalization weight vectors and the final model normalization weights. The aggregate ratio is $\sum_s\widehat V_{\mathrm{MGPO},s}/\sum_s\widehat V_{\mathrm{Factual},s}$. Its 95\% percentile confidence interval uses 5,000 paired document-cluster bootstrap resamples, retaining all states per sampled document and holding their estimated variances fixed. 

\subsection{Prompt Settings}
\label{app:prompts}

We report abridged prompts that preserve the task definition, model inputs, and evaluation-relevant constraints. Full runtime prompts are released with the implementation.

Across tasks, the writer uses the same update interface. Given a task specification, previous memory, and current chunk, it produces the next memory state. Tasks differ only in the specification supplied to the writer and the downstream readout.

\subsubsection{Task Specifications}

The task specification defines the downstream objective without prescribing how information should be represented in memory.

\textbf{SciREX.}
The schema contains four entity types (Method, Metric, Task, and Material). Salient entities participate in reported experimental results, and relations connect entity pairs participating in the same result.

\textbf{AIPAN-10K.}
The schema contains five entity types (Data type, Data Purpose, Recipient, Data Handling, and User Right). Relations connect each data type to its stated privacy practices.

\textbf{BABILong.}
The writer receives the sample question as the task specification. Benchmark-specific instructions and demonstrations are provided only to the reader.

\subsubsection{Memory Writer Prompt}
\label{app:memory-writer-prompts}

All tasks use the same memory-update interface. At each step, the writer receives the previous memory and the current document chunk together with the task specification.

\begin{promptbox}{Memory writer prompt}
Follow the task below. The previous memory and current chunk are provided as inputs.
<task>{task}</task>
You are presented with a task, a chunk of the document that may contain information relevant to the task, and a previous memory. Read the chunk carefully and update the memory with new information that helps future predictions. Return only the updated memory.
<memory>{memory}</memory>
<chunk>{chunk}</chunk>

Updated memory:
\end{promptbox}

For SciREX and AIPAN-10K, \texttt{task} contains the corresponding extraction task description above. For BABILong, it only contains the benchmark question: \texttt{question}. The initial memory is \texttt{No previous memory}, and each update replaces the preceding memory. No memory schema or examples are prescribed. The memory budget is enforced during generation.

\subsubsection{Reader Prompts}
\label{app:reader-prompts}

\paragraph{IE Readout.}
For SciREX and AIPAN-10K, the frozen reader combines the current chunk with memory and returns structured extraction predictions:

\begin{promptbox}{IE Reader prompt}
You are an information-extraction system. Return exactly one JSON object.
<task>
{task_description}
{extraction_instructions}
</task>
<memory>{memory}</memory>
<chunk>{chunk}</chunk>

Extraction:
\end{promptbox}

For SciREX, the reader extracts typed entities, groups aliases, predicts salience, and returns unordered relations between salient entities of different types participating in the same result:
\texttt{{entities: [{id, name, type, mentions, salient}], relations: [{head, tail, type}]}}.

For AIPAN-10K, the reader extracts minimal verbatim mentions of data types and associated privacy practices, omits explicitly denied practices, and returns:
\texttt{{entities: [{id, name, type, mentions}], relations: [{head, tail}]}}.

\paragraph{BABILong Readout.}

After the final chunk, the reader receives only the final memory and the question, using the benchmark's released instruction, demonstrations, and answer-format prompt:

\begin{promptbox}{BABILong reader prompt}
System: You are a helpful assistant.
{instruction}
{examples}
{post\_prompt}
<context>{final_memory}</context>

Question: {question}
\end{promptbox}

The fields \texttt{examples} and \texttt{post\_prompt} contain the benchmark's task-specific demonstrations and answer-format requirement, respectively. The original document chunks are not provided at readout. 

\section{Additional Experimental Results}

\subsection{Complete SciREX Results}
\label{app:scirex-span}
For compatibility with prior SciREX work \citep{jain2020scirex,huang2021document}, we additionally report binary-relation scores using the unmodified TempGen evaluation script. This evaluator first maps predicted clusters to gold clusters by span overlap and then scores relations over the mapped identities, making it relatively insensitive to cluster fragmentation. We therefore use our structured IE scorer as the primary measure of entity-cluster recovery. Table~\ref{tab:full_scirex} also reports typed-mention extraction as a diagnostic. MGPO retains 44.3 F1 despite receiving no reward on this metric, whereas R1-RE, MemAgent, and HiMPO obtain 29.4, 23.5, and 24.2 F1, respectively, consistent with reduced interference when memory learning is decoupled from the frozen reader.

\begin{table*}[h]
\caption{Complete SciREX test results. \emph{Structured IE} uses our document-level scorer, while \emph{SciREX} reports binary relations using the unmodified TempGen evaluator \citep{huang2021document}. All non-prior-work results are mean@4.}

\label{tab:full_scirex}
\centering
\small
\setlength{\tabcolsep}{3pt}
\renewcommand{\arraystretch}{0.94}
\begin{tabular}{l ccc ccc ccc ccc}
\toprule
& \multicolumn{9}{c}{Structured IE} & \multicolumn{3}{c}{SciREX} \\
\cmidrule(lr){2-10}\cmidrule(lr){11-13}
& \multicolumn{3}{c}{Typed Mention} & \multicolumn{3}{c}{Salient cluster}
& \multicolumn{3}{c}{Binary relation} & \multicolumn{3}{c}{Binary relation} \\
\cmidrule(lr){2-4}\cmidrule(lr){5-7}\cmidrule(lr){8-10}\cmidrule(lr){11-13}
Method & P & R & F1 & P & R & F1 & P & R & F1 & P & R & F1 \\
\midrule
\multicolumn{13}{l}{\textit{Prior work}} \\
SciREX-P & 50.3 & 90.2 &\textbf{64.6} & 17.6 & 78.4 & 28.8 & 2.8 & 58.2 & 5.4 & 6.5 & 41.1 & 9.6 \\
TempGen  & 79.0 & 7.9 & 14.4 & 24.2 & 23.7 & 23.9 &8.3 & 6.3 & 7.1 & 17.1 & 13.6 & 14.5 \\
\midrule
\multicolumn{13}{l}{\textit{Direct Readout}} \\
Direct Readout                     & 65.9 & 25.3 & 36.5 & 29.5 & 58.4 & 39.2 & 10.8 & 35.0 & 16.5 & 9.8 & 22.0 & 11.8 \\
R1-RE                          & 72.6 & 18.4 & 29.4 & 43.9 & 59.8 & 50.6 & 19.2 & 24.9 & 21.7 & 19.5 & 24.9 & 20.1 \\
\midrule
\multicolumn{13}{l}{\textit{External Memory}} \\
LightRAG                 & 58.4 & 43.4 & 49.8 & 27.4 & 58.9 & 37.4 & 10.2 & 32.0 & 15.4 & 14.3 & 29.1 & 16.8 \\
Mem0                     & 50.0 & 45.7 & 47.7 & 34.6 & 66.9 & 45.6 & 15.8 & 41.6 & 22.9 & 16.8 & 31.8 & 19.8 \\
\midrule
\multicolumn{13}{l}{\textit{Learned Memory}} \\
MemAgent                       & 86.3 & 13.6 & 23.5 & 49.8 & 50.1 & 49.9 & 22.6 & 16.7 & 19.2 & 21.5 & 22.0 & \textbf{20.9} \\
HiMPO                   & 72.0 & 14.6 & 24.2 & 39.8 & 47.6 & 43.3 & 15.4 & 17.7 & 16.5 & 20.8 & 23.3 & 20.5 \\
\midrule
\multicolumn{13}{l}{\textit{Decoupled Memory}} \\
MGPO (ours)      & 56.4 & 36.4 & 44.3 & 44.2 & 61.0 & \underline{\textbf{51.3}} & 20.8 & 36.8 & \underline{\textbf{26.5}} & 19.8 & 27.2 & 20.8 \\
\bottomrule
\end{tabular}
\end{table*}

\subsection{BABILong Results}
\label{app:babilong}
\begin{figure}[h]
    \centering
    \includegraphics[width=1\linewidth]{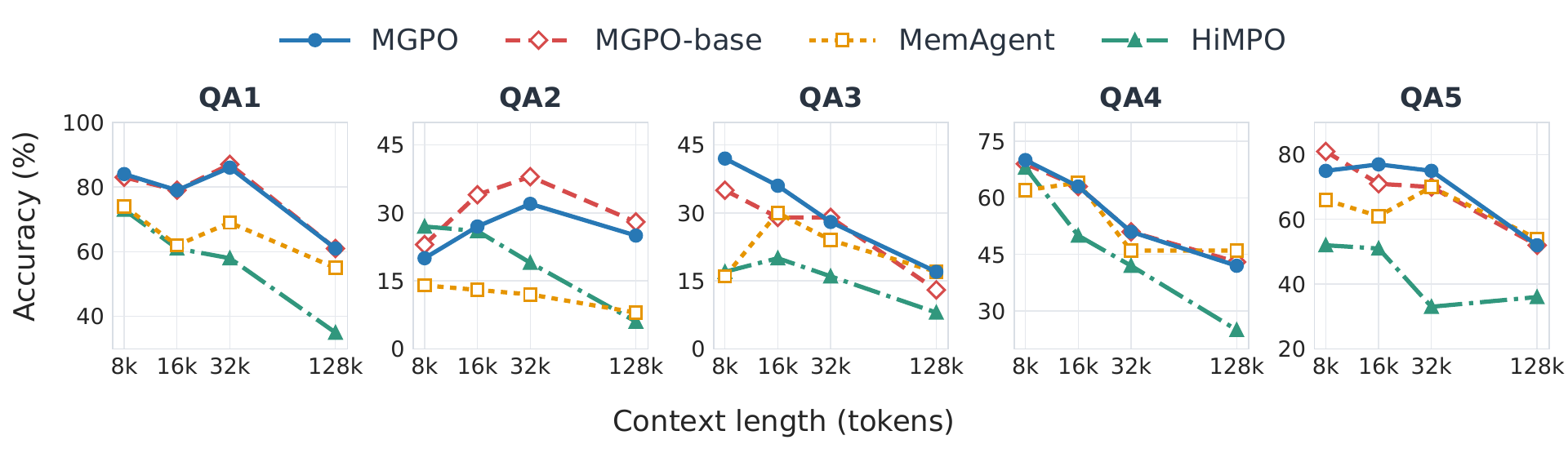}
    \caption{Accuracy on BABILong QA1–QA5 across context lengths.}
    \label{fig:babilong}
\end{figure}

As shown in \autoref{fig:babilong}, MGPO remains effective across diverse QA tasks and context lengths up to 128K, consistently outperforming HiMPO and MemAgent in most settings and matching or exceeding MGPO-base on several tasks, particularly QA3 and QA5. These results indicate that MGPO learns a reusable strategy for preserving downstream-relevant information that transfers across downstream tasks and output schemas.

\section{Mechanism and Diagnostic Analyses}
\subsection{Empirical Fidelity of the Surrogate}
\label{app:linearized_plot}
\begin{figure}[H]
\centering
\includegraphics[width=0.75\linewidth]{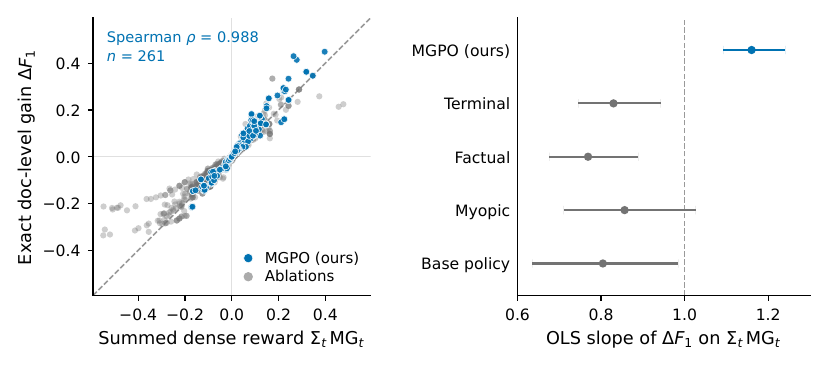}
\caption{
\textbf{The linearized reward closely tracks exact F1 gain.}
\textbf{Left:} Exact document-level F1 gain versus summed Memory Gain on SciREX, computed from the same history-masked counterfactual counts. MGPO (blue; $n=261$) achieves Spearman $\rho=0.988$ and matching signs whenever both gains are nonzero; credit ablations and the base policy are shown in grey.
\textbf{Right:} OLS slopes of exact on linearized gain with document-bootstrap 95\% CIs ($B=2000$). The surrogate slightly underestimates MGPO gains (1.16, 95\% CI [1.09, 1.24]) and overestimates ablation loss magnitudes (0.77--0.86).
}
\label{fig:linearization}
\end{figure}

\subsection{Robustness to Reader Sampling}
\label{app:reader_noise}
\begin{figure}[h]
\centering
\includegraphics[width=0.4\textwidth]{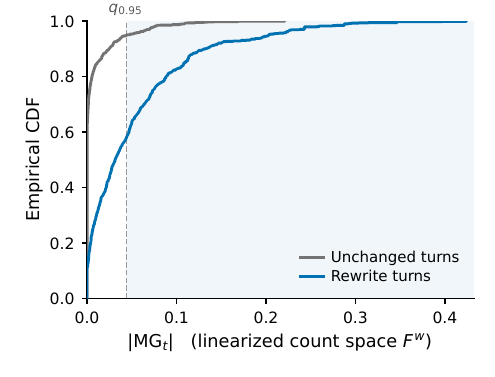}
\caption{
\textbf{Noise in single-sample utility estimation.}
Empirical CDFs of $|MG_t|$ for unchanged-memory controls (grey) and actual rewrites (blue). The dashed line is the 95th percentile of the control distribution.
}
\label{fig:reader_noise}
\end{figure}

We use temperature-based decoding because greedy decoding substantially degrades Qwen3 extraction quality, and estimate each $F_{t,j}$ from one independently sampled reader output. To quantify the resulting sampling noise, we evaluate unchanged-memory trajectories across eight random seeds as an empirical noise reference. As shown in \autoref{fig:reader_noise}, 85\% of the total absolute rewrite-credit mass comes from rewrites whose Memory Gain exceeds the 95th percentile of this reference, indicating that most credit lies well above single-sample variation.

\subsection{Memory-Budget and Chunk-Size Sensitivity}
\begin{table}[H]
\centering
\small
\caption{Sensitivity to memory budget and chunk size. We conduct experiments using the model trained with a configuration of a 1024 tokens chunk size and a 256 tokens memory budget. Performance is largely insensitive to the memory budget, while varying the chunk size shows a minor effect, peaking at 2048. $^{\dagger}$ denotes the training configuration.}
\label{tab:geometry}
\begin{tabular}{rr cc}
\toprule
Memory & Chunk & Salient cluster & Binary relation \\
\midrule
\multicolumn{4}{l}{\textit{Memory budget, chunk fixed at 1024}} \\
 64 & 1024 & 50.8 {\scriptsize $\pm$ 0.4} & 26.1 {\scriptsize $\pm$ 0.7} \\
128 & 1024 & 51.3 {\scriptsize $\pm$ 1.2} & 26.4 {\scriptsize $\pm$ 1.6} \\
256$^{\dagger}$ & 1024$^{\dagger}$ & 51.3 {\scriptsize $\pm$ 0.9} & 26.5 {\scriptsize $\pm$ 1.3} \\
512 & 1024 &  51.5 {\scriptsize $\pm$ 0.7} & 26.7 {\scriptsize $\pm$ 0.8} \\
\midrule
\multicolumn{4}{l}{\textit{Chunk size, memory fixed at 256}} \\
256 &  512 &  47.9 {\scriptsize $\pm$ 0.7} & 24.8 {\scriptsize $\pm$ 0.4} \\
256$^{\dagger}$ & 1024$^{\dagger}$ &  51.3 {\scriptsize $\pm$ 0.9} & 26.5 {\scriptsize $\pm$ 1.3} \\
256 & 2048 & 51.2 {\scriptsize $\pm$ 0.5} & 28.0 {\scriptsize $\pm$ 0.6} \\
256 & 4096 & 49.8 {\scriptsize $\pm$ 0.5} & 25.3 {\scriptsize $\pm$ 0.5} \\
\bottomrule
\end{tabular}
\end{table}

\subsection{Position-EMA Estimator}

We compare our position-specific EMA baseline with a group-relative alternative while keeping the Memory Gain objective unchanged. To accommodate variable document lengths, we maintain position-specific baselines only up to a cutoff $K$, and pool all later rewrites into a shared tail bucket, i.e., $b(t)=b(K)$ for $t\ge K$. This avoids increasingly sparse position estimates for long documents while retaining position-dependent calibration over the earlier trajectory. As shown in \autoref{tab:estimator-group}, Position EMA performs better on the document-level tasks than group-relative method while requiring only one sample per state.
\begin{table*}[h]
\centering
\small
\caption{
\textbf{Position EMA versus group-relative baseline on SciREX.}
Both variants use the same Memory Gain objective. MGPO-EMA uses position EMA ($G=1$), whereas MGPO-GR uses GRPO ($G=8$).
Policy rollouts and reader evaluations are reported per training document.
GPU-hours cover the complete training run, and peak memory is the
analytical per-step training footprint.
}
\label{tab:estimator-group}
\setlength{\tabcolsep}{3.5pt}
\begin{tabular}{lcrrrrrrr}
\toprule
Method
& $G$
& \shortstack{Mem.\\tokens}
& \shortstack{Policy\\rollouts}
& \shortstack{Reader\\evaluations}
& \shortstack{GPU-\\hours}
& \shortstack{Peak mem.\\(GiB)}
& Cluster F1
& Relation F1 \\
\midrule
MGPO-base
& --
& 202
& --
& --
& --
& --
& 37.8 $\pm$ 0.4
& 15.5 $\pm$ 0.4 \\

MGPO-GR
& 8
& 72
& 8
& 50.7
& 142.0
& 198.4
& 45.5 $\pm$ 0.5
& 20.1 $\pm$ 0.3 \\

MGPO-EMA
& 1
& 43
& 1
& 8.3
& 26.9
& 192.5
& \textbf{51.3} $\pm$ 0.9
& \textbf{26.5} $\pm$ 1.3 \\
\bottomrule
\end{tabular}
\end{table*}

\subsection{Case Study}
\label{sec:case-study}
We inspect two points along a SciREX evaluation trajectory from the SciREX-trained MGPO writer. Chunk indices are zero-based. Each box shows source excerpts and the complete memory after that chunk; \texttt{[...]} denotes omitted source text. Line breaks are for layout.

The seven-chunk paper ``Training Region-based Object Detectors with Online Hard Example Mining'' introduces OHEM and its evaluation datasets in chunk~0. The writer retains a compact result record while later chunks describe the training algorithm and implementation details.

\begin{promptbox}{SciREX: chunk 0 and memory 0}
Chunk 0 (excerpts; [...] marks omissions)
[...]
We present a simple yet surprisingly effective online hard example mining ( OHEM ) algorithm for training region - based ConvNet detectors .
[...]
Moreover , combined with complementary advances in the field , OHEM leads to state - of - the - art results of 78.9\% and 76.3\% mAP on PASCAL VOC 2007 and 2012 respectively .
[...]

Memory 0 
{
  "Method": "OHEM",
  "Metric": "mAP",
  "Task": "Object detection",
  "Material": "PASCAL VOC 2007",
  "Metric-Task": "mAP-Object detection",
  "Method-Metric": "OHEM-mAP",
  "Method-Task": "OHEM-Object detection",
  "Method-Material": "OHEM-PASCAL VOC 2007",
  "Task-Material": "Object detection-PASCAL VOC 2007",
  "Metric-Material": "mAP-PASCAL VOC 2007"
}
\end{promptbox}

Chunks~1--3 leave the memory unchanged. Chunk~3 mentions OHEM and Caffe,
but contains no gold alias of PASCAL VOC 2007. Caffe appears only as the
implementation framework.

\begin{promptbox}{SciREX: chunk 3 and memory 3}
Chunk 3 (excerpts; [...] marks omissions)
[...]
More specifically , the online hard example mining algorithm ( OHEM ) proceeds as follows .
[...]
We implement both options described above using the Caffe framework ( see ) .
[...]

Memory 3
{
  "Method": "OHEM",
  "Metric": "mAP",
  "Task": "Object detection",
  "Material": "PASCAL VOC 2007",
  "Metric-Task": "mAP-Object detection",
  "Method-Metric": "OHEM-mAP",
  "Method-Task": "OHEM-Object detection",
  "Method-Material": "OHEM-PASCAL VOC 2007",
  "Task-Material": "Object detection-PASCAL VOC 2007",
  "Metric-Material": "mAP-PASCAL VOC 2007"
}
\end{promptbox}

At chunk~3, the memory-conditioned reader predicts
\texttt{OHEM}--\texttt{PASCAL VOC 2007}, combining the locally observed
method with the dataset retained in memory. Without memory, the reader instead
predicts two relations to \texttt{Caffe framework}, neither matching a gold
relation. The retained result context thus separates evaluation material from
implementation context across chunks.

\section{Computational Analysis}
\subsection{Computation Complexity}
\label{app:complexity}
\begin{figure} [H]
    \centering
    \includegraphics[width=0.9\linewidth]{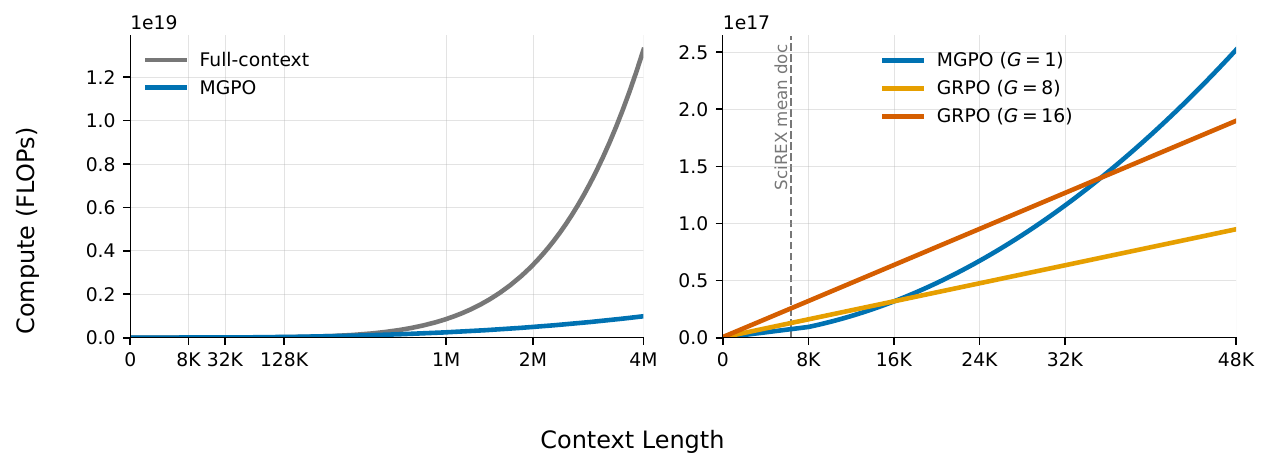}
    \caption{
        Analytical FLOPs under the SciREX instrument, every call charged at its configured maximum width. \textbf{Left: }Inference: a full-context 14B reader scales quadratically in document length while chunk-wide MGPO scales linearly. \textbf{Right: }Training: Memory Gain (G=1) pays a triangular counterfactual reward matrix, GRPO pays G trajectories with diagonal reader cells. At the mean SciREX (train) document (dashed line, 6,346 tokens) Memory Gain costs half of GRPO-8, and only overtakes it past 16K tokens.}
    \label{fig:comp_flops}
\end{figure}

\paragraph{Practical training cost.}
During training, we use only chunks containing attributed gold structures as reader-evaluation targets in the counterfactual matrix. Chunks without attributed targets are still processed by the memory policy, so their rewrites remain part of the trajectory and can affect utility on subsequent supervised targets, but they are not themselves scored as target chunks. This sparse target evaluation substantially reduces the number of reader calls relative to the dense estimate in \autoref{fig:comp_flops}. Our transfer experiments further show that memory policies trained on shorter documents remain effective on substantially longer contexts, reducing the need to train directly at the longest deployment lengths.

\subsection{GPU Usage}
\begin{figure} [H]
    \centering
    \includegraphics[width=0.9\linewidth]{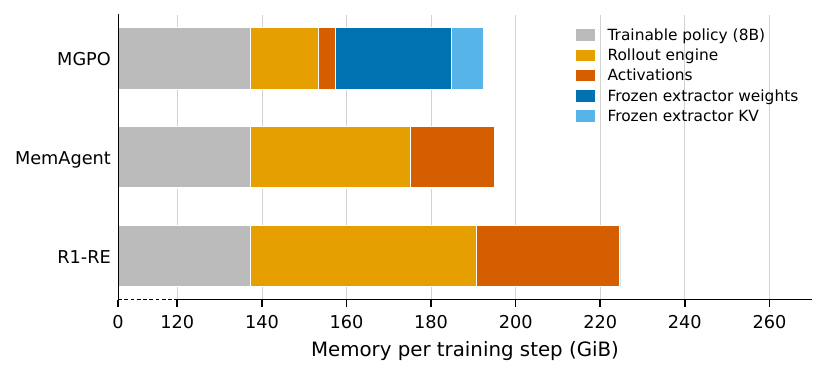}
   \caption{
            \textbf{Training-time GPU memory breakdown on SciREX.}
            All methods train the same Qwen3-8B policy under their respective training configurations. Policy memory is therefore identical, while rollout KV cache and activations vary with each method's sequence and batch geometry. MGPO additionally maintains a frozen Qwen3-14B reader and its KV cache. Nevertheless, its chunk-wise training setup substantially reduces rollout and activation memory, yielding a total footprint comparable to MemAgent and lower than R1-RE.
            }
    \label{fig:comp_mem}
\end{figure}

\section{Scope and Limitations}
MGPO isolates the marginal effect introduced between adjacent memory states, but does not explicitly decompose higher-order interactions among multiple rewrites. When downstream utility emerges only after several complementary updates, the newly realized gain is attributed to the rewrite at which that utility becomes observable rather than divided among the preceding contributing updates. This does not eliminate long-horizon credit to earlier actions, since subsequent Memory Gains remain included in their cumulative returns \(G_t\). Rather, it limits the granularity of rewrite-level attribution. Explicit interaction-aware or coalitional attribution is an interesting direction for future work.

\end{document}